\RequirePackage{fix-cm}
\documentclass[twocolumn]{svjour3}
\smartqed

\usepackage{graphicx}
\usepackage{amsmath,amssymb}
\usepackage{bm}
\usepackage{booktabs}
\usepackage{multirow}
\usepackage{upgreek}
\usepackage{url}
\usepackage{color,xcolor}
\usepackage{bbding}
\usepackage[colorlinks,linkcolor=blue,anchorcolor=blue,citecolor=blue,urlcolor=blue]{hyperref}
\usepackage{xurl}   
\usepackage{mathptmx}

\emergencystretch=\maxdimen
\newcommand{\rawvol}{raw volume}
\newcommand{\featvol}{feature volume}
\newcommand{\dsf}{s}                     
\newcommand{\rEPE}{\mbox{$EPE_{\mathrm{raw}}$}}

\newcommand{\Rraw}{R_{\mathrm{raw}}}       
\newcommand{\Rfeat}{R_{\mathrm{feature}}}  
\newcommand{\Uraw}{U_{\mathrm{raw}}}       
\newcommand{\Ufeat}{U_{\mathrm{feature}}}  
\newcommand{\ourmethod}{RAFT-DVC}    

\journalname{Acta Mech. Sin.}

\begin{document}

\title{RAFT-DVC: Resolution-Aware Machine Learning-Based Digital Volume Correlation  %
}
\titlerunning{RAFT-DVC: a resolution-aware framework for ML-based DVC}

\author{Zixiang Tong$^1$ \and Lehu Bu$^2$ \and Jin Yang$^{1,2}$ {\Envelope}}
\authorrunning{Z.~Tong et al.}

\institute{$^1$ Department of Aerospace Engineering and Engineering Mechanics,
           The University of Texas at Austin, Austin, TX 78712, USA   \\
           $^2$ Materials Science Graduate Program, Texas Materials Institute,
           The University of Texas at Austin, Austin, TX 78712, USA \\
           {\Envelope} Corresponding author, \email{jin.yang@austin.utexas.edu}}

\date{
}

\maketitle

\begin{abstract}
Digital volume correlation (DVC) provides three-dimensional full-field displacement measurements from volumetric images, but how the internal resolution of a machine-learning-based DVC model affects accuracy and operating range remains poorly understood. Here, we present RAFT-DVC, a resolution-aware family of recurrent all-pairs field transforms (RAFT)-based DVC solvers with encoder downsampling factors $s$ = 2, 4, and 8. Using a matched design, we find that the three solvers localize displacement to approximately 0.017 feature-grid voxels, giving an empirical raw-volume error scaling of approximately 0.017$s$ voxel.

The solvers exhibit complementary operating regimes governed jointly by displacement reach and volumetric-texture compatibility. Synthetic benchmarks show that RAFT-DVC achieves errors of the same order as tuned classical DVC under fine-texture, small-to-moderate-displacement conditions and becomes competitive or advantageous under coarse-texture, large-displacement conditions. Frequency-swept tests quantify deformation spatial resolution, while tiled inference enables dense estimation on large volumes. Evaluation on confocal volumetric images acquired during indentation illustrates the importance of matching solver operating regime to deformation magnitude and image texture. Tests on micro-CT images of elastomeric foam, despite training only on particle-labeled synthetic data, provide evidence of cross-texture transfer.

We also identify coordinate-order inconsistencies in three-dimensional RAFT correlation sampling and introduce a non-cubic impulse test to verify sampler geometry independently of network training. Correcting the sampler improves native-input accuracy and generalization to unseen volume dimensions. Together, these results establish RAFT-DVC as a fast, resolution-aware framework for dense DVC with characterized accuracy and operating regimes.

\keywords{Digital volume correlation \and Machine learning \and Optical flow
\and Experimental mechanics }
\end{abstract}

\section{Introduction}
\label{sec:intro}
Digital volume correlation (DVC) measures three-dimensional displacement and strain fields by matching reference and deformed image volumes~\cite{bay1999dvc,bay2008methods}. Because DVC resolves deformation within a specimen rather than only on its surface, it has become an important tool for quantifying internal deformation and mechanics across a wide range of materials and imaging modalities~\cite{tong2026dvcchallenge}. Applications include granular and cellular solids imaged by X-ray computed tomography~\cite{hu2014internal,landauer_materials_2023}, soft and biological tissue imaged by confocal or multiphoton microscopy~\cite{franck2007confocal,maskarinec2009quantifying,stout2016pnas,lavigne2022digital}, bone and other biological tissues imaged by optical coherence tomography (OCT)~\cite{midgett2019oct} or magnetic resonance imaging (MRI)~\cite{benoit20093d}, and geomaterials imaged by neutron tomography~\cite{tudisco2015full}.

To date, classical DVC post-processing analysis algorithms can be broadly classified as local \cite{bay1999dvc,bay2008methods,tudisco2017tomowarp2,barkochba2015fidvc}, global \cite{van2019global,buljac2018digital,leclerc2011voxel}, or hybrid \cite{yang2020aldvc} methods. Among them, local DVC methods remain widely used because of their conceptual simplicity and flexibility. Local DVC methods divide the reference volume into subvolumes and estimate displacement independently at a set of measurement
points, typically located at the subvolume centers. Displacement may be estimated from the peak of a spatial- or Fourier-domain cross-correlation function~\cite{barkochba2015fidvc,chen1993digital,jiang2015path}. Iterative
methods extend this formulation through repeated image warping and refinement; for example, Fast Iterative DVC (FIDVC)~\cite{barkochba2015fidvc} uses iterative updates to improve displacement estimation. Alternatively, local DVC can minimize a grayscale-intensity residual under a prescribed displacement shape function. Combined with inverse-compositional Gauss--Newton (IC-GN) optimization, this approach can recover sub-voxel displacement and local displacement gradients with high accuracy
~\cite{tudisco2017tomowarp2,baker2004lucas,gates2011towards,gao_high_efficiency_2015}.

Besides local methods, global DVC methods instead represent the displacement field using a set of basis functions and determine all degrees of freedom through volume-wide optimization. Finite-element-based global formulations represent the displacement field on a finite element mesh and can incorporate spatial regularization to reduce sensitivity to image noise~\cite{van2019global,leclerc2011voxel,hild2016toward,wittevrongel2015self,passieux2019classic}. In addition, dense optical-flow formulations similarly estimate the full-field displacement vector at every voxel by combining image matching with spatial regularization. Global DVC methods provide kinematically compatible displacement fields, although their overall performance depends on factors including spatial discretization, regularization,  initialization, and the validity of the underlying image-intensity conservation assumption.

More recently, hybrid local-global methods seek to combine advantages of the two approaches.
For example, augmented-Lagrangian DVC (ALDVC)~\cite{yang2020aldvc} couples
locally evaluated correlation problems to a globally compatible displacement
field, thereby combining local correlation with global kinematic consistency
~\cite{yang2019augmented,yang2019combining}.

\subsection{Current Challenges}
Despite substantial progress, several challenges continue to affect the accuracy, robustness, and computational efficiency of conventional DVC methods~\cite{buljac2018digital,pan2020advances}. First, DVC performance depends on user-selected analysis parameters, including subvolume or finite-element size, regularization strength, search range, initialization strategy, and convergence criteria. These parameters often need to be adjusted when the imaging condition, volumetric texture, or displacement magnitude changes. For instance, a subvolume that is too small may contain insufficient image information for robust correlation, whereas an excessively large subvolume may reduce spatial resolution and smooth local deformation gradients.


Second, DVC depends intrinsically on the volumetric texture provided by the specimen and the used imaging modality. Unlike 2D DIC~\cite{yang2021smart,sugerman2023speckling} or 3D stereo DIC~\cite{tong20253d}, for which an artificial speckle pattern can often be applied, the interior of a specimen generally cannot be textured directly. Volumes acquired by X-ray computed tomography, confocal microscopy, optical coherence tomography, magnetic resonance imaging, and other modalities may therefore contain substantially different microstructural features, characteristic feature sizes, spatial densities, contrast levels, and noise characteristics~\cite{tong2026dvcchallenge}. Sparse, blurred, or spatially nonuniform texture can reduce correlation quality, particularly when the characteristic image features are poorly matched to the spatial scale of the measurement. Consequently, DVC does not generally admit a single optimal set of analysis parameters for all specimens and experiments; instead, the measurement configuration must be
adapted to the available image texture,  deformation regime, and specific type of research problems~\cite{bay2008methods}.

Third, the computational cost of conventional DVC increases rapidly with volume size and measurement spatial resolution~\cite{leclerc2012digital,gates2015high}. Local DVC methods perform a cross-correlation or an iterative optimization at every measurement point, whereas global DVC methods repeatedly evaluate and solve a coupled optimization problem over the entire volume domain. Depending on the image dimensions, node spacing, and solver settings, processing a three-dimensional image pair can require minutes to hours even when the resulting displacement field is sampled substantially more coarsely than the voxel grid. This cost becomes increasingly restrictive for high-resolution and time-resolved experiments that contain hundreds of volume pairs and terabyte-scale datasets.

\subsection{Recent Advances in Machine Learning-Based DVC Methods and Their Limitations}

Recent advances in machine learning have motivated new DVC methods that replace or reformulate parts of the conventional optimization procedure. These methods can be broadly organized into two computational paradigms. 
The first comprises \emph{pretrained predictors}, which are trained on a collection of volume pairs and subsequently estimate displacement through a forward pass. Within this paradigm, some networks directly regress displacement from image data, whereas others explicitly establish correspondence between learned representations. The second paradigm comprises \emph{instance-optimized neural solvers}, in which a neural representation is fitted separately to each reference--deformed volume pair. This distinction is important because the two paradigms have different advantages and limitations: pretrained predictors amortize the computational cost of learning across many subsequent measurements but inherit a dependence on the training distribution, whereas instance-optimized methods avoid a fixed training database but retain an optimization procedure for every analyzed volume pair.

Among pretrained approaches, DVC-Net~\cite{duan2022deepdvc}, following earlier
direct-regression optical-flow networks~\cite{dosovitskiy2015flownet,sun2018pwc}, was one of the earliest dedicated deep-learning frameworks for three-dimensional DVC. It uses three convolutional subnetworks to sequentially perform integer-voxel displacement estimation, sub-voxel registration, and displacement-field denoising. By replacing conventional correlation searches and iterative refinement with learned operations, DVC-Net demonstrated that DVC analysis could be accelerated by two to three orders of magnitude relative to conventional methods. Its quantitative evaluation, however, was primarily based on synthetically generated textures and prescribed deformation fields over a limited range, while the experimental examples lacked independently known displacement fields. Consequently, performance outside the represented texture and deformation distributions, as
well as metrological characteristics such as zero-strain precision, spatial
resolution, and operating range, remained incompletely characterized.

VolRAFT~\cite{wong2024volraft} introduced a correspondence-based alternative by extending recurrent all-pairs field transforms (RAFT)~\cite{teed2020raft} from two-dimensional optical flow to volumetric displacement measurement. Specifically, VolRAFT extracts learned representations from the reference and deformed volumes, constructs a six-dimensional all-pairs correlation volume, and iteratively refines a dense displacement field using a recurrent update operator. This formulation provides an explicit learned correspondence mechanism and was demonstrated on synchrotron micro-CT images of bone--implant interfaces. As a three-dimensional extension of the original RAFT architecture, however, VolRAFT retains an encoder downsampling factor of $\dsf{}=8$. Displacement is therefore refined on an internal feature grid whose spacing is eight raw voxels and subsequently interpolated to the raw-volume grid. Such a coarse internal grid provides a large nominal correlation reach in raw-volume coordinates but raises an important question for experimental mechanics: how does feature-grid spacing affect the sub-voxel displacement accuracy ultimately reported on the measurement grid? Related work in learning-based DIC has similarly highlighted
the importance of internal feature resolution for high-accuracy deformation
measurement~\cite{pan2024raftdic,tong2026raftcorr}. In three dimensions, the all-pairs
correlation formulation also introduces a steep memory cost that limits direct
processing of large volumes.

Transformer-based methods have also been extended from 2D DIC to DVC. The DICTr framework formulates deformation measurement as learned feature matching using a transformer encoder~\cite{zhou2025dictr}, while its unsupervised volumetric extension, DVCTr~\cite{he2026dvctr}, estimates displacement by matching learned features between reference and deformed volumes at multiple resolutions. Rather than requiring prescribed displacement fields during training, unsupervised DVCTr warps one volume using the estimated displacement and minimizes the resulting image mismatch. A multi-resolution displacement-gradient consistency term is introduced to reduce noise-induced variations without imposing a conventional smoothness penalty. DVCTr demonstrated promising performance on simulated and experimental traction-force-microscopy volumes. Its training objective, however, still depends on image warping and approximate correspondence of material features between the two volumes, and the recovered solution may depend on the relative weighting of the image-matching and gradient-consistency terms. More generally,
its zero-strain precision, spatial resolution, displacement operating range, and transfer across substantially different volumetric textures remain to be systematically characterized.

A complementary class of methods uses neural networks as instance-optimized, continuous representations of the displacement field. PiNetDVC~\cite{wang2026pinetdvc} takes spatial coordinates as input and represents displacement and strain as continuous neural functions. The network is fitted independently to each reference--deformed volume pair by combining image consistency with a strain--displacement compatibility constraint evaluated through automatic differentiation. This formulation does not require a labeled training database and avoids explicitly discretizing the field into independent subsets or finite elements. It also permits displacement and strain to be obtained within a unified representation. However, the network must be optimized separately for each volume pair, so the computational cost is relocated from conventional DVC optimizations to neural-network optimization rather than amortized through pretraining.  

Moreover, the spectral bias and inherent continuity of conventional coordinate networks can limit their ability to represent displacement discontinuities and high-spatial-frequency deformation. RFPI-DVC~\cite{wang2026random} extends PiNetDVC by introducing random Fourier feature mapping into the coordinate-based representation. The Fourier mapping is intended to reduce spectral bias and improve the reconstruction of high-frequency deformation features, including shear bands, crack-tip fields, and steep strain gradients. The method was evaluated using numerical examples and \emph{in situ} X-ray CT measurements of triaxially compressed sand. Nevertheless, like other instance-optimized approaches, it
requires optimization and hyperparameter selection for each volume pair, and
its performance across other imaging modalities, texture scales, and
deformation regimes remains to be established.

Taken together, existing machine learning-based DVC methods address some limitations of conventional DVC methods. DVC-Net demonstrated rapid learned displacement inference; VolRAFT introduced recurrent all-pairs correspondence in three dimensions; DVCTr reduced the requirement for labeled displacement fields; PiNetDVC and RFPI-DVC introduced continuous, mechanically constrained neural representations. 
A remaining challenge, however, is how to \emph{characterize and select} a pretrained DVC solver for a particular measurement condition. In correspondence-based networks, the relationship between internal feature-grid spacing and raw-volume displacement accuracy has
not been systematically isolated. Likewise, the coupled effects of feature-grid
resolution, volumetric texture, displacement reach, spatial resolution, computational cost, and out-of-distribution performance have not been mapped in a common experimental framework.

\subsection{Contributions of this work}

To address these gaps, we develop \ourmethod{}, a resolution-aware family of RAFT-based DVC solvers for volumetric displacement measurement. The framework is systematically trained and characterized using synthetic particle-labeled volumes, with additional experimental evaluations used to assess transfer beyond the training texture. The framework uses three encoder downsampling factors, $\dsf{}\in\{2,4,8\}$, enabling controlled characterization of how internal feature-grid spacing influences displacement accuracy and how trained solvers operate across different texture and deformation regimes.

The main contributions of this work are as follows.

First, we introduce the RAFT-DVC framework in Section~\ref{sec:arch} that consists of three solvers with different encoder downsampling factors ($\dsf{}=2$, 4, and 8).  We characterize their displacement and texture operating regimes. Rather than seeking a single universally optimal network, the framework treats solver selection as a measurement-design choice that depends on the volumetric texture and deformation regime.

Second, using a matched feature-grid design, we establish an empirical relationship between internal feature-grid spacing and raw-volume displacement accuracy. By scaling particle geometry, particle density, displacement magnitude, and input-volume dimensions such that the three solvers encounter comparable problems in feature-grid coordinates, we show that the displacement error remains approximately constant in feature-grid units and consequently
scales with $\dsf{}$ when expressed in raw-volume voxels (see Sections~\ref{sec:ruler}-\ref{sec:solver}).

Third, we introduce a coordinate-consistency test for three-dimensional RAFT correlation sampling and quantify the effect of sampler consistency on displacement accuracy and generalization across volume dimensions (see Sections~\ref{sec:sampler} and \ref{sec:sizegen_results}). This provides a network-independent diagnostic for validating the geometric implementation of three-dimensional correlation lookup.

Finally, we systematically evaluate \ourmethod{} against representative classical local, global, and augmented-Lagrangian DVC methods and against VolRAFT in terms of displacement accuracy, zero-strain noise floor, spatial resolution, out-of-distribution behavior, computational cost, and large-volume deployment (see Sections~\ref{sec:noisefloor}-\ref{sec:baseline}). We further evaluate the trained solvers on an experimental particle-labeled confocal volume (see Section~\ref{sec:realdata}) and release the implementation, trained models, sampling diagnostic, and synthetic-data generation tools to support reproducibility.

\begin{figure*}[t]
  \centering
  \includegraphics[width=\textwidth]{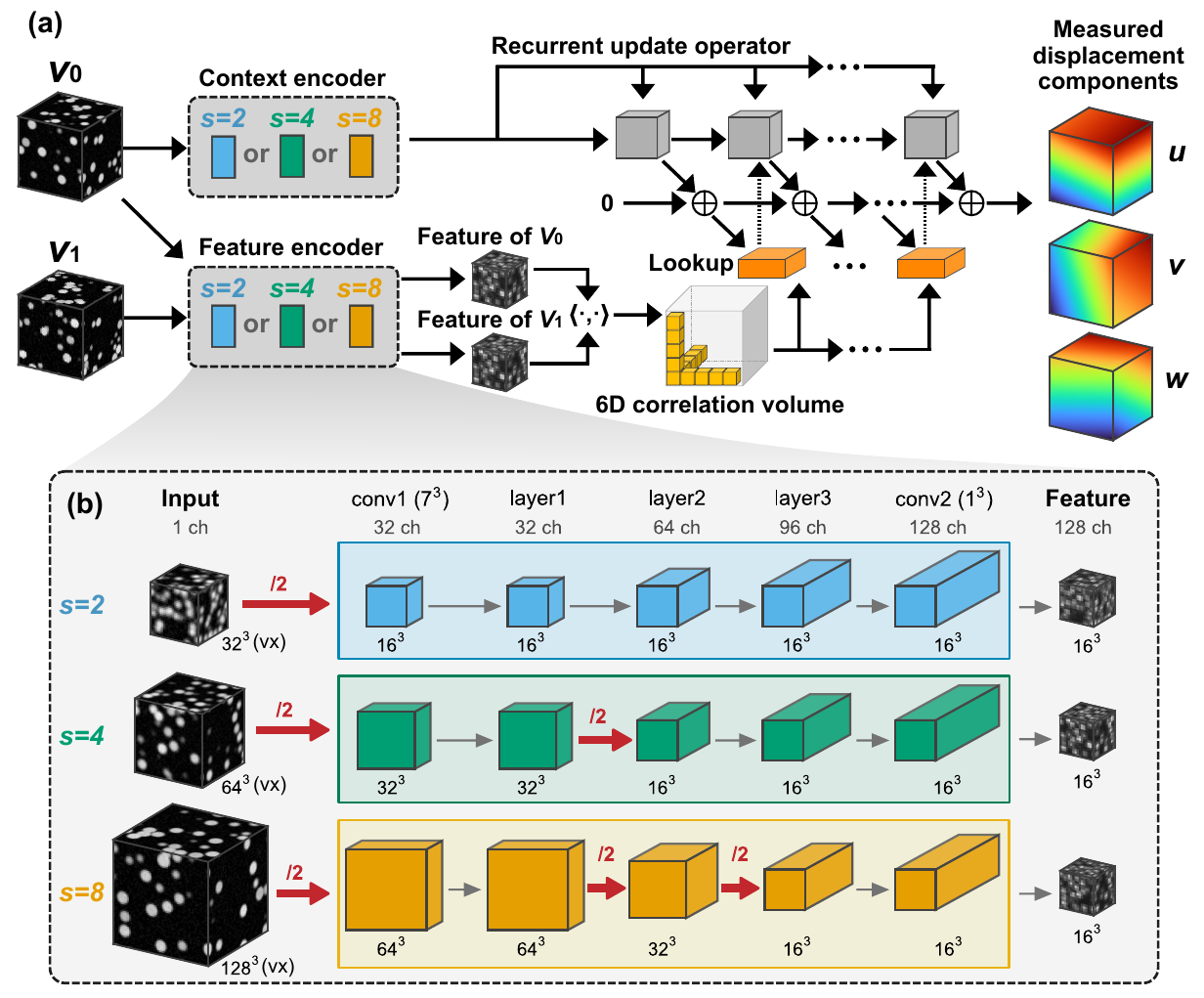}
  \caption{\ourmethod{} architecture and encoder configurations.
\textbf{(a)} Overall RAFT-DVC workflow. The reference and deformed volumes, $V_0$ and $V_1$, are processed by a shared feature encoder, and pairwise inner products between the resulting \featvol{} representations form an all-pairs six-dimensional correlation volume. A context encoder applied to $V_0$ initializes the recurrent update operator. At each iteration, correlation features are sampled around the current displacement estimate (orange rectangles), and the recurrent operator outputs a displacement increment that is accumulated ($\oplus$) into the evolving three-component field $(u,v,w)$.
\textbf{(b)} Feature-encoder configurations for the s2 ($s = 2$), s4 ($s = 4$), and s8 ($s = 8$) arms. The three encoder variants use the same convolutional and residual block types and channel widths but differ in the placement of stride-2 downsampling operations (red arrows, ``/2''). For the matched design illustrated here, input volumes of $32^3$, $64^3$, and $128^3$ voxels for s2, s4, and s8, respectively, are mapped to the same $16^3$ \featvol{} grid. Spatial dimensions are indicated beneath each intermediate representation.}
  \label{fig:overview}
\end{figure*}

\section{Methodology}
\label{sec:method}
\subsection{RAFT-DVC Architecture}
\label{sec:arch}

\ourmethod{} takes a reference volume $V_0$ and a deformed volume $V_1$ as input and returns a dense voxel-wise three-dimensional displacement field on the \rawvol{} grid (see Fig.~\ref{fig:overview}(a)). The architecture follows the recurrent all-pairs correlation formulation of RAFT~\cite{teed2020raft}, extended here to volumetric displacement estimation. Table~\ref{tab:conceptmap} summarizes functional analogies between key components of \ourmethod{} and familiar elements of conventional DVC methods. 

A shared three-dimensional feature encoder, with weights shared between the two inputs, maps $V_0$ and $V_1$ from the \rawvol{} grid to 128-channel \featvol{} representations.
As illustrated in Fig.~\ref{fig:overview}(b), the feature encoder consists of a $7 \times 7 \times 7$ convolutional stem, followed by residual stages with channel widths $32 \rightarrow 32 \rightarrow 64 \rightarrow 96$, and a final $1\times1\times1$ projection to 128 channels. Instance normalization without affine parameters is applied within the feature encoder. A separate context encoder operates only on reference image $V_0$ (see Fig.\,\ref{fig:overview}(a)``Feature encoder''). It uses the same backbone structure but without normalization and provides a 96-channel hidden state together with 64 context
channels to initialize the recurrent update operator.

\begin{table}[h!]
\centering
\caption{Conceptual correspondence between components of \ourmethod{} and
familiar elements of classical DVC. The pairings indicate functional analogies
rather than one-to-one equivalences; in particular, learned feature
representations and training distributions do not have unique classical
counterparts. }
\label{tab:conceptmap}
\begin{tabular}{@{}p{0.45\linewidth}p{0.52\linewidth}@{}}
\toprule
\ourmethod{} concept & Classical DVC method analogue \\
\midrule
Feature-grid spacing $\dsf{}$ & Subvolume size or finite-element discretization scale \\
Learned feature representation & Raw-image intensity texture  \\
Correlation radius and pyramid levels & Search range and multiscale search strategy \\
Estimated displacement increment   & Incremental displacement correction \\
Convolutional GRU update      & Iterative displacement refinement \\
Training distribution         & Prior assumptions and parameter selection\\
Trilinear upsampling          & Displacement-field interpolation \\
Tiled inference               & Domain decomposition \\
Overlapping-tile blending     & Overlap stitching / partition-of-unity blending \\
\bottomrule
\end{tabular}
\end{table}

An all-pairs 3D correlation volume is constructed from the inner products between
the two \featvol{}s. The correlation volume is average-pooled into a pyramid of
$L$ levels, each level halving the grid of the one before it, and we use $L=2$.
At each recurrent iteration, a separable 3D convolutional gated recurrent unit
(ConvGRU) queries a $(2r+1)^3$ neighborhood centered on the current displacement
estimate at every pyramid level (see Fig.\,\ref{fig:overview}(a)``Recurrent
update operator''), where $r$ is the search radius in \featvol{} voxels; we use
$r=4$, so the queried neighborhood is $9^3$. The ConvGRU outputs a
three-component displacement increment.

The displacement is further refined on the \featvol{} grid and subsequently mapped to the \rawvol{} grid using trilinear interpolation. Unlike the original RAFT architecture, we do not use a learned convex upsampling operator. The complete network of RAFT-DVC contains about $2.12$ million trainable parameters.


\subsection{Encoder Arms and Matched Controlled Design}
\label{sec:disp-design}
\begin{figure*}[t]
  \centering
  \includegraphics[width=0.82\textwidth]{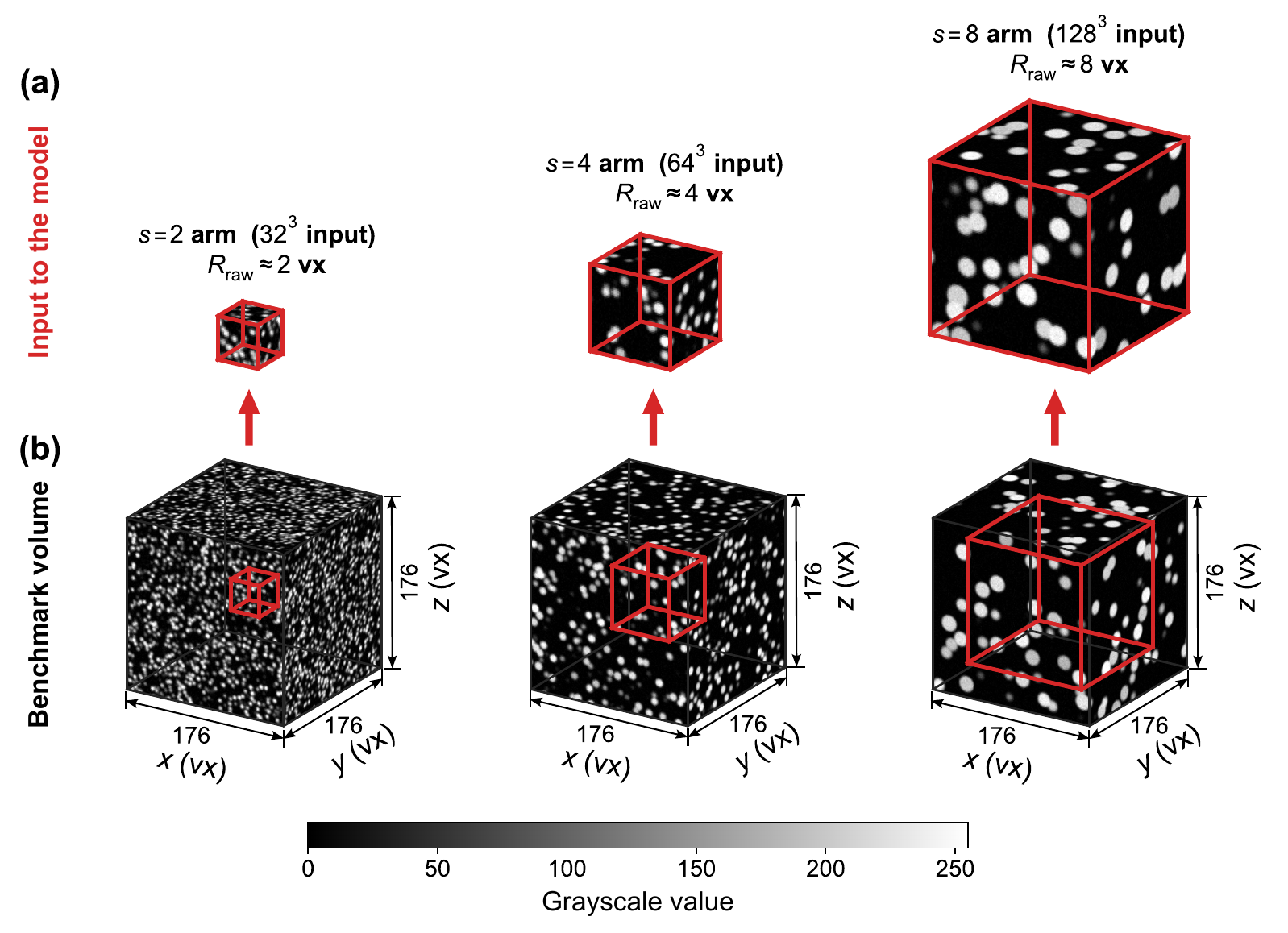}
  \caption{Synthetic particle volumes and matched input scaling for the three \ourmethod{} arms. \textbf{(a)} Representative model inputs for s2, s4, and s8, with raw-volume sizes of $32^3$, $64^3$, and $128^3$ voxels and particle radii $\Rraw{}=2$, 4, and 8 voxels, respectively. Both the input dimensions and particle radius scale with the encoder downsampling factor $\dsf{}$, such that the particle radius is identical in feature-grid coordinates, $\Rfeat{}=\Rraw{}/\dsf{}=1$ feature voxel, for all three arms.
\textbf{(b)} Representative $176^3$-voxel synthetic benchmark volumes for the corresponding fine-, intermediate-, and coarse-particle regimes.Red boxes show, at a common display scale, the spatial extent corresponding to the model input of each arm. All volumes are displayed using the same grayscale range. The $176^3$ benchmark volumes are generated independently at their native size; the red boxes indicate model-input extent only and do not denote cropped subvolumes.}
  \label{fig:data3d}
\end{figure*}
%
For clarity, we distinguish three related terms used throughout this paper. An \emph{encoder variant} refers to a particular stride configuration of the feature and context encoders, which sets the encoder downsampling factor $\dsf{}$. An \emph{arm} comprises an encoder variant together with the particle-size and displacement distribution on which it is trained and is the unit compared in the controlled experiments below. A \emph{solver} refers to a trained arm used for inference on a measurement. 

We consider three arms, denoted s2, s4, and s8, with different encoder downsampling factors $\dsf{}\in\{{2,4,8}\}$, respectively. The three encoder variants contain the same convolutional and residual stages, channel dimensions, and number of trainable parameters, but differ in the placement of stride-2 downsampling operations (``/2'') (see Fig.~\ref{fig:overview}b). The spatial grid is reduced once along each axis
for s2, twice for s4, and three times for s8. Because the final channel dimensions are unchanged, all three encoder variants contain the same number of trainable parameters.

The downsampling factor, $s$, determines the spacing of the internal \featvol{} grid on which displacement is iteratively refined. One step on the \featvol{} grid therefore corresponds to $\dsf{}$ voxels along each spatial axis of the \rawvol{}. Accordingly, the s2, s4, and s8 arms have feature-grid spacings of 2, 4, and 8 raw voxels, respectively. Under the matched design described below, all three arms operate on the same $16^3$ \featvol{} grid; what differs is the physical extent represented by each feature-grid voxel.

Every spatial quantity in this study can therefore be expressed either in \rawvol{} coordinates or in \featvol{} coordinates. The corresponding length scales differ by the downsampling factor $\dsf{}$. In particular, we define
\begin{align}
    \Rfeat{} &= \frac{\Rraw{}}{\dsf{}}, \\
     \Ufeat{} &= \frac{\Uraw{}}{\dsf{}}, 
\end{align}
where $\Rraw{}$ and $\Rfeat{}=\Rraw{}/\dsf{}$ denote particle radius measured in \rawvol{}  and \featvol{} voxels, respectively; $\Uraw{}$ and $\Ufeat{}$ denote the
corresponding displacement component magnitudes such that $\max_{\mathbf{x},i} |{u}_i(\mathbf{x})| = U_{\mathrm{raw}}$.

To examine the effect of feature-grid spacing while keeping the internal correlation problem comparable across the three arms, we construct a \emph{matched feature-grid design}. The input-volume size, particle radius, particle number density, and displacement magnitude are scaled with $\dsf{}$ such that all three arms operate on the same $16^3$ \featvol{} grid and encounter the same particle geometry, particle density, displacement range, and feature-grid dimensions when these quantities are expressed in feature-volume coordinates (see Fig.~\ref{fig:data3d}). 

Specifically, s2, s4, and s8 arms are trained using raw input volumes of $32^3$, $64^3$, and $128^3$ voxels, respectively. Their corresponding particle radii are $ \Rraw{} = 2,\ 4,\ 8$ voxels, so that all three arms have the same particle radius in feature-grid coordinates, $\Rfeat{}=1$ feature voxel. The corresponding particle number densities are $4.8$, $0.6$, and $0.075$ particles per $10^3$ \rawvol{} voxels, respectively. Because one \featvol{} voxel corresponds to $\dsf{}^3$ \rawvol{} voxels, these values are equivalent to the same particle density of  38.4 particles per $10^3$~\featvol{} voxels for all three arms.
Similarly, the training displacement ranges are $\Uraw{}$ $\in$ $\{[2,4]$, [4,8], [8,16]$\}$~voxels for s2, s4, and s8, respectively. These ranges correspond to the same
 interval in feature-grid coordinates, i.e.,  $\Ufeat{}\in[1,2]$ feature voxels.

This construction therefore matches the particle geometry, particle density, displacement range, and correlation-grid dimensions in feature-volume coordinates while varying the raw-volume spacing represented by one feature-grid voxel. It provides a controlled basis for examining how encoder downsampling is associated with displacement accuracy in \rawvol{} coordinates. Importantly, this comparison is not a fixed-input encoder ablation. If the same
raw particle volume and displacement field were supplied to encoders with different $\dsf{}$, both particle size and displacement magnitude would change simultaneously when expressed on the feature grid. Such fixed-raw-volume, cross-regime conditions are therefore examined separately as out-of-distribution (OOD) tests.

\subsection{Synthetic Particle-Volume Generation}
\label{sec:data}
Training and validation are performed using synthetic particle-labeled volumes with prescribed ground-truth displacement fields. All volumes are generated using a deterministic pipeline with
configuration-hash seeding to ensure reproducibility. Particles are rendered as spheres of uniform interior intensity with a one-voxel-thick linear edge taper centered on the nominal particle radius. The
edge taper provides a continuous variation of grayscale intensity with sub-voxel particle position, avoiding the discontinuous changes that would arise from rasterizing binary spheres directly onto the voxel grid. Overlapping particles are combined by intensity saturation, after which the complete volume is convolved with an isotropic Gaussian point-spread function with standard deviation $\sigma=0.8$ voxel.

Particle radius and intensity are independently perturbed to introduce intra-volume variability. For particle $n$, the radius is sampled as $R_n \sim \mathcal{U} \left(\Rraw{}-0.2,\,\Rraw{}+0.2\right) $ voxels, where $\mathcal{U}(\bullet)$ means a uniform distribution; $\Rraw{}$ is the nominal particle radius for the corresponding configuration. The normalized particle intensity is independently sampled from a Gaussian distribution $\mathcal{N}(1, 0.01)$, where a nominal particle has unit intensity.

Each synthetic reference--deformed volume pair is associated with a prescribed three-dimensional displacement field expressed as a polynomial in the voxel coordinates. The polynomial expansion is taken about the volume center coordinates $\mathbf{x}_c$. Defining $\Updelta\mathbf{x}=\mathbf{x}-\mathbf{x}_c$, the $i$-th displacement component is written as
\begin{equation}
  u_i(\mathbf{x}) \;=\; p^{(0)}_i
  \;+\; \sum_{j=1}^{3} p^{(1)}_{ij}\,\Delta x_j
  \;+\; \sum_{j=1}^{3}\sum_{k=j}^{3} p_{ijk}^{(2)}\,\Delta x_j\Delta x_k ,
  \label{eq:shapefn}
\end{equation}
where $i,j,k\in\{1,2,3\}$ denote the $x$, $y$, and $z$ directions, such that $(u_1,u_2,u_3)$ are the displacement components $(u,v,w)$. The coefficients $p^{(0)}_i$, $p^{(1)}_{ij}$, and $p^{(2)}_{ijk}$ are the constant, first-order, and second-order polynomial coefficients. Truncating Eq~\eqref{eq:shapefn} at zeroth, first, or second order produces rigid translation, affine, or quadratic displacement fields. These three field classes are sampled with probabilities of $10\%$, $45\%$, and $45\%$, respectively. Before amplitude normalization, all active polynomial coefficients are sampled independently from a normal distribution $\mathcal{N}(0,1)$. The same sampling procedure is applied independently to the three displacement components, so no spatial direction is preferentially represented.

Because Eq~\eqref{eq:shapefn} is linear in its coefficients, the spatial form and overall amplitude of each sampled displacement field can be controlled independently. After a field is generated, all coefficients are multiplied by a single common scale factor such that $\max_{\mathbf{x},\,i}     \left|u_i(\mathbf{x})\right| = \Uraw{}$,
over the evaluated region, where the target displacement magnitude
$\Uraw{}$ is sampled uniformly from the range assigned to that configuration.

In this paper, the deformed volume is generated directly from displaced particle coordinates rather than by interpolating the rendered reference volume, which is closer to real confocal microscopy imaging experiments.  Specifically, the center of particle $n$ in the reference volume, $\mathbf{x}_n$, is displaced to $\mathbf{x}'_n=\mathbf{x}_n+\mathbf{u}(\mathbf{x}_n)$, with Eq~\eqref{eq:shapefn} evaluated at that particle's continuous, generally off-grid position. The reference and deformed volumes are then rendered independently from the particle-center sets $\{\mathbf{x}_n\}$ and $\{\mathbf{x}'_n\}$ using the same particle renderer and point-spread function. Particles are initially generated within an expanded
domain, and both rendered volumes are subsequently center-cropped to the target dimensions so that particle motion does not introduce particles into or remove particles from the field of view.

Imaging noise is added independently to each rendered volume. Each voxel first receives Poisson shot noise followed by additive Gaussian read noise, $\xi$,
\begin{equation}
  \tilde{I} \;=\; \frac{1}{\kappa}\,\mathrm{Poisson}\!\big(\kappa I\big)
  \;+\; \xi , \qquad \xi\sim\mathcal{N}\!\big(0,\,\sigma_{\mathrm{read}}^{2}\big),
  \label{eq:noise}
\end{equation}
where $I$ is the noise-free normalized intensity; $\kappa=500$ is the photon count per unit intensity, and $\sigma_{\mathrm{read}}=0.01$. Negative grayscale values are clipped to zero. The Poisson contribution therefore varies with the local signal level, whereas the read-noise variance is spatially uniform.

The same generation procedure is used for all three \ourmethod{} arms; only the matched particle radius, particle density, input-volume size, and displacement range differ according to Section~\ref{sec:disp-design}. For each arm, we generate 2,000 training volume pairs and 200 validation pairs. Complete
generation and seeding parameters are provided in Appendix~\ref{app:disclosure}.

\subsection{Training}
\label{sec:training}
All three RAFT-DVC arms are trained for 300 epochs using the same optimization procedure and an effective batch size of 8. Gradient accumulation is used when necessary to maintain the same effective batch size across arms. The learning-rate schedule is indexed by optimizer updates rather than individual micro-batches.

Optimization is performed using AdamW~\cite{loshchilov2017decoupled} with a weight decay of $5\times10^{-5}$. We employ a one-cycle learning-rate schedule with a warm-up fraction of 0.2, a peak learning rate of $2\times10^{-4}$ followed by cosine annealing after the warm-up phase. Gradients are clipped to a maximum $\ell_2$ norm of $1.0$, and training is performed using mixed-precision (fp16) computation~\cite{micikevicius2017mixed}. Unless otherwise stated, all training runs use random seed 42 with deterministically seeded data generation and augmentation.

Following the original RAFT formulation~\cite{teed2020raft}, supervision is applied to the displacement estimate at every recurrent iteration. For $N=12$ recurrent updates, the sequence loss is
\begin{equation}
  \mathcal{L} = \sum_{i=0}^{N-1} \gamma^{\,N-1-i}\,
  \frac{1}{N_{vx}} \sum_{ \mathbf{x} \in \Upomega_{\mathrm{training}} }
  \left\| \hat{\mathbf u}_i(\mathbf{x})-\mathbf u_{\mathrm{GT}}(\mathbf{x}) \right\|_1,
  \label{eq:raft_loss}
\end{equation}
where $\hat{\mathbf{u}}_i(\mathbf{x})$ denotes the estimated displacement at voxel $\mathbf{x}$ after the $i$th recurrent update; $\mathbf{u}_{\mathrm{GT}}(\mathbf{x})$ is the prescribed ground-truth displacement; $N_{vx}$ is the number of voxels in the training volume $\Upomega_{\mathrm{training}}$; and $\gamma$ controls the exponential weighting of intermediate estimates. We use $\gamma=0.8$, such that the final estimate receives unit weight and progressively earlier estimates receive smaller weights.

\subsection{Evaluation Metrics}
\label{sec:metrics}
%
\subsubsection*{Displacement Accuracy}

Displacement accuracy is quantified using the endpoint error (EPE), reported in both \rawvol{} and \featvol{} coordinates.  The raw-volume endpoint error, $EPE_{\mathrm{raw}}$, is defined as the mean Euclidean distance between the estimated displacement field $\hat{\mathbf{u}}$ and the prescribed ground-truth displacement field $\mathbf{u}_{\mathrm{GT}}$,
\begin{equation}
    EPE_{\mathrm{raw}} = \frac{1}{N_{vx}} \sum_{ \mathbf{x} \in \Upomega_{\mathrm{evaluation}} } \left\| \hat{\mathbf{u}}(\mathbf{x}) - \mathbf{u}_\mathrm{GT}(\mathbf{x}) \right\|_{2}, \label{eq:epe_raw}
\end{equation} 
where $\Upomega_{\mathrm{eval}}$ denotes the evaluated region and $N_{\mathrm{vox}}$ is the number of voxels within that region. $EPE_{\mathrm{raw}}$ is expressed in \rawvol{} voxels and therefore quantifies displacement error directly in the coordinate system of the measurement volume. 

To compare displacement localization across different encoder downsampling factors, we also report the same error in feature-volume coordinates, 
\begin{equation}
     EPE_{\mathrm{feature}} = \frac{EPE_{\mathrm{raw}}}{\dsf{}}, \label{eq:epe_feature}
\end{equation}
which is expressed in feature-volume voxels. Thus, $EPE_{\mathrm{raw}}$ and $EPE_{\mathrm{feature}}$ represent the same
displacement error expressed on the two spatial grids. 

We also highlight that our training objective uses an $\ell_1$ sequence loss (see Section~\ref{sec:training}), whereas evaluation uses the Euclidean $\ell_2$ endpoint error, which is consistent with common practice in optical-flow and other learning-based DVC benchmarking~\cite{wong2024volraft,teed2020raft,butler2012naturalistic}.

\subsubsection*{Reference-free Image-matching Residual}
Evaluation using EPE requires a known ground-truth displacement field and therefore cannot be applied directly to experimental measurements. For such data, we complement displacement-field comparisons with a reference-free metric that quantifies how well a resolved displacement field aligns the reference and deformed image volumes. Following the squared intensity residual introduced for learned two-dimensional correlation by Liu et al.~\cite{liu2026sir} and building on the gray-level residual formulation of Chi et al.~\cite{chi2023glr}, we extend the metric here to volumetric data.

Let $V_0(\mathbf{x})$ and $V_1(\mathbf{x})$ denote the reference and deformed volumes, respectively, and let $\hat{\mathbf{u}}(\mathbf{x})$ denote the measured displacement field. Both volumes are first low-pass filtered using an isotropic Gaussian kernel with standard deviation $\sigma_d$, yielding $V_{0d}$ and $V_{1d}$. This preprocessing reduces the contribution of high-spatial-frequency acquisition noise to the image-matching residual. We use $\sigma_d=1$ voxel and represent image grayscale intensities on a 0--255 grayscale range throughout this study. 

Assuming that $\hat{\mathbf{u}}(\mathbf{x})$ maps a material point at reference coordinate $\mathbf{x}$ to its location in the deformed volume, the filtered deformed volume is sampled back at the corresponding reference coordinate as
\begin{equation}
    V_{1w}(\mathbf{x}) = V_{1d} \left(\mathbf{x}+\hat{\mathbf{u}}(\mathbf{x})  \right),
    \label{eq:volume_warp}
\end{equation}
where $V_{1d}$ is evaluated at non-integer voxel coordinates by tricubic
B-spline interpolation, matching the bicubic interpolation conventionally used
in the classical solvers we compare against.\\

To account for local brightness and contrast differences between reference and deformed volume acquisitions, including depth-dependent attenuation in confocal image stacks, we apply a local affine intensity correction. For a cubic window
$\Upomega_{\mathbf{x}}$ of side length $w$ centered at $\mathbf{x}$, the local
contrast coefficient $a(\mathbf{x})$ and brightness offset $b(\mathbf{x})$
are determined by minimizing the squared intensity mismatch. The resulting
local mean squared intensity residual is
\begin{equation}
    C_{\mathrm{SSD}}(\mathbf{x}) = \frac{1}{|\Upomega_{\mathbf{x}}|}  \sum_{\mathbf{x}_i \in \Upomega_{\mathbf{x}}}
    \left[ a(\mathbf{x})V_{0d}(\mathbf{x}_i)  + b(\mathbf{x}) - V_{1w}(\mathbf{x}_i) \right]^2 ,
    \label{eq:SSD}
\end{equation}
where $|\Upomega_{\mathbf{x}}|$ is the number of voxels in the local window. We use $w=21$ voxels throughout this study.

The minimizing coefficients of each window $\Upomega$ are obtained analytically as
\begin{align}
\left\lbrace
\begin{aligned}
    a(\mathbf{x})
    &=  \frac{ \mathrm{cov}_{\Upomega_{\mathbf{x}}}  \left(V_{0d},V_{1w}\right) }{
        \operatorname{var}_{\Upomega_{\mathbf{x}}}  \left(V_{0d}\right) },    \\
    b(\mathbf{x})
    &=  \left\langle V_{1w}\right\rangle_{\Upomega_{\mathbf{x}}} - a(\mathbf{x}) \left\langle V_{0d}\right\rangle_{\Upomega_{\mathbf{x}}},
    \end{aligned}
    \right.
\end{align}
where $\langle\cdot\rangle_{\Upomega_{\mathbf{x}}}$ denotes the local mean, and $\operatorname{cov}_{\Upomega_{\mathbf{x}}}(\cdot,\cdot)$ and $\operatorname{var}_{\Upomega_{\mathbf{x}}}(\cdot)$ denote the local covariance and variance
computed over $\Upomega_{\mathbf{x}}$, respectively.

One important thing to note is that $C_{\mathrm{SSD}}$ quantifies local image-matching quality rather than displacement accuracy. A displacement field that overfits image noise may produce a small residual without being mechanically accurate. We therefore use $C_{\mathrm{SSD}}$ only as a complementary, reference-free validation measure and interpret it together with the agreement between \ourmethod{} and classical DVC methods. As a baseline, the measured residual is also compared with that obtained from the null displacement field, $\hat{\mathbf{u}}_0(\mathbf{x})\equiv\mathbf{0}$.

\subsubsection*{Deformation Spatial Resolution}
Displacement accuracy and image-matching quality quantify how accurately a
solver recovers a displacement field, but they do not characterize the spatial
scale of deformation that can be resolved. Like conventional correlation-based
methods, a DVC solver attenuates displacement variations at sufficiently short
spatial wavelengths~\cite{yang2021fast}. For a prescribed sinusoidal displacement field, this attenuation appears as a reduction in recovered amplitude as the deformation wavelength decreases~\cite{reu2022dic}. We therefore define the \emph{attenuation curve} as the ratio of recovered to prescribed displacement amplitude as a function of deformation wavelength. The \emph{critical wavelength} $\lambda_q$ is defined as the wavelength at which the recovered amplitude reaches a fraction $q$ of the prescribed amplitude; for example, $\lambda_{85}$ denotes the wavelength at which $85\%$ of the prescribed amplitude is recovered.

The attenuation curve is measured from a frequency-swept sinusoidal displacement field described in Section~\ref{sec:baseline}, following the DIC
Challenge convention~\cite{reu2018dicchallenge}. The
recovered displacement profile is extracted along the sweep direction and averaged over the transverse
direction along which the prescribed displacement is spatially uniform. At each
local maximum and minimum of the prescribed profile, the retained amplitude
fraction is evaluated as $u_\mathrm{pred}/u_\mathrm{GT}$,
where $u_\mathrm{pred}$ and $u_\mathrm{GT}$ denote the recovered and 
prescribed displacements, respectively, at the corresponding extremum. Both
maxima and minima are included, providing two measurements per deformation
cycle. 

The retained amplitude fraction is fitted as a function of position along the
frequency sweep using a decreasing sigmoid,
\begin{equation}
  \frac{u_\mathrm{pred}}{u_\mathrm{GT}}(x)
  = \frac{T_{0}}{1+\exp\!\left[k\,(x-x_{0})\right]},
  \label{eq:sigmoid}
\end{equation}
where $x$ denotes position along the sweep direction, $T_{0}$ is the
long-wavelength plateau, $k>0$ controls the transition width, and $x_{0}$
denotes the transition center. The fit is performed in position coordinates
because the prescribed wavelength varies deterministically with position; this
avoids estimating a local wavelength from the spacing between neighboring
extrema. After fitting, the position at which the attenuation curve reaches a
specified fraction $q$ is mapped to the corresponding prescribed wavelength to
obtain $\lambda_q$.

The lower asymptote of the sigmoid is fixed at zero, corresponding to complete
attenuation in the limit of vanishing deformation wavelength. A sigmoid is used
instead of the polynomial fit employed in the DIC Challenge
~\cite{reu2018dicchallenge} because it enforces a monotonic attenuation curve,
provides an explicit long-wavelength plateau, and yields a unique crossing for
each specified attenuation level.

\begin{table*}[h!]
\centering
\caption{Training cost and all-pairs correlation-memory requirements of the three \ourmethod{} arms, measured on an NVIDIA RTX~5090 GPU with 32~GB of memory. Training uses mixed-precision computation (fp16) and is performed for 300 epochs with an effective batch size of 8. The left block reports the raw input size, micro-batch and gradient-accumulation configuration, wall-clock time per epoch, total training time, and measured peak GPU memory. The right block reports the fp32 memory required by the all-pairs correlation volume during direct inference as a function of raw
input size and encoder downsampling factor. For a cubic \featvol{} with side length $n$, the all-pairs correlation volume contains $\mathcal{O}(n^6)$ entries. ``Tiled'' indicates that direct inference exceeds the intended memory budget and overlapping tiled inference is used instead; ``--'' denotes configurations not evaluated.}
\label{tab:cost}
\begin{tabular}[t]{cccccc}
\toprule
\multicolumn{6}{c}{\textbf{Training cost}} \\ \midrule
Arm & Raw imput size (vx) & Micro-batch $\times$ accum. & Time/epoch & Total time & Peak GPU memory \\
\midrule
s2 & $32^3$  & $8\times1$ & $69$~s  & $5.7$\,h & $18.6$\,GB \\
s4 & $64^3$  & $8\times1$ & $76$~s  & $6.3$\,h & $19.2$\,GB \\
s8 & $128^3$ & $4\times2$ & $176$~s & $14.7$\,h & $13.2$\,GB \\
\bottomrule \\
\end{tabular}

\hspace{2.5em}
\begin{tabular}[t]{cccc}
\toprule
\multicolumn{4}{c}{\textbf{All-pairs correlation memory during inference}} \\
\midrule
Raw input size (vx) & s2 & s4 & s8 \\
\midrule
$32^3$  & 0.07~GB & --      & --      \\
$64^3$  & 4.3~GB  & 0.07~GB & --      \\
$80^3$  & 16.4~GB & 0.16~GB & --      \\
$128^3$ & Tiled   & 4.3~GB  & 0.07~GB \\
\bottomrule
\end{tabular}
\end{table*}

\subsection{Correlation-Sampler Validation and Tiled Inference} 
\label{sec:sampler}  
%
The three-dimensional RAFT correlation lookup requires consistent conventions for the memory layout of the correlation volume, the ordering of spatial coordinates in the sampling grid, and the normalization applied along each axis. Extending the original two-dimensional RAFT formulation to three dimensions introduces an additional spatial axis and therefore increases the risk of silent coordinate-order inconsistencies. Such inconsistencies can be difficult to detect when both training and inference are performed on fixed-size cubic volumes, because tensor dimensions remain compatible and the network may partially adapt to the resulting correlation representation.

To verify the sampling geometry independently of network training, we use a non-cubic impulse test. A unit impulse is placed at a prescribed location in a non-cubic three-dimensional volume and queried using the same sampling operator employed by the correlation lookup. A coordinate-consistent sampler must return the impulse at the prescribed location. Because the three spatial dimensions are unequal, an axis permutation produces an immediately detectable shift in the recovered impulse location. We additionally verify the coordinate mapping using non-cubic volumes populated with randomly generated feature values.

This validation procedure revealed coordinate-order inconsistencies in an early implementation of our three-dimensional sampler and, independently, in the released VolRAFT implementation~\cite{wong2024volraft}. We corrected the sampling convention so that the correlation-volume layout, sampling-grid coordinates, and axis-wise normalization use a consistent spatial ordering. All \ourmethod{} results reported in this study use the corrected sampler and were obtained after retraining the networks. For comparison, VolRAFT is retrained and evaluated using both its released sampler and a
coordinate-corrected version; the corresponding quantitative results are reported in Section~\ref{sec:sizegen_results} and Section~\ref{sec:baseline}.

Direct inference on large volumes is constrained by the memory requirement of the three-dimensional all-pairs correlation volume. For a cubic \featvol{} with side length $n$, the correlation volume stores a similarity value for every pair of feature-grid locations and therefore contains
$\mathcal{O}(n^6)$ entries. Volumes that exceed the available direct-inference memory budget are processed using overlapping cubic tiles. Each tile is evaluated independently, and displacement estimates in overlapping regions are combined to reduce boundary artifacts. The implementation details of the blending procedure are described in Section~\ref{sec:cost}, while the effects of input size, tile size, and overlap on displacement accuracy are quantified in Section~\ref{sec:sizegen_results}.

\subsection{Computational Cost and Memory Requirements}
\label{sec:cost}
Training and primary inference benchmarks are performed on a single NVIDIA RTX~5090 GPU with 32~GB memory using mixed-precision fp16 computation, while the all-pairs correlation volume and correlation lookup are stored and evaluated in fp32. The multi-seed experiments described in Section~\ref{sec:ruler} are trained on a Texas Advanced Computing Center (TACC) Vista GH200 node. Complete hardware and software specifications are provided in Appendix~\ref{app:disclosure}.

The dominant memory cost of RAFT-based three-dimensional correlation is the all-pairs correlation volume. For a cubic \featvol{} with side length $n$, there are $n^3$ feature locations in each input volume, and a correlation value is stored for every pair of locations. The resulting correlation volume therefore contains $(n^3)^2=n^6$ entries, so its memory requirement grows with the sixth power of the feature-grid side length. For a fixed raw-volume size, reducing the encoder downsampling factor $\dsf{}$ increases the feature-grid dimensions and therefore increases correlation memory rapidly.

Under the matched feature-grid design of Section~\ref{sec:disp-design}, all three arms operate on the same $16^3$ \featvol{} grid during training and therefore have the same correlation-volume size. Differences in measured peak training memory therefore reflect other components of the computation, particularly encoder activations and the micro-batch configuration. To maintain a common effective batch size of 8 within the available GPU memory, s2 and s4 use micro-batches of 8, whereas s8 uses a micro-batch of 4 with two-step gradient accumulation (Table~\ref{tab:cost}).

For raw volumes that exceed the practical memory limit of direct inference, we use overlapping tiled inference as introduced in Section~\ref{sec:sampler}. The volume is partitioned into cubic tiles with 50\% overlap, each tile is processed independently, and predictions in overlapping regions are blended using a normalized raised-cosine (Hann) window~\cite{harris1978use}. For a fixed tile size, this strategy makes peak GPU memory depend primarily on the tile dimensions rather than on the dimensions of the complete volume, thereby allowing inference on substantially larger datasets. The effects of tile size and overlap on displacement accuracy are quantified in Section~\ref{sec:sizegen_results}.

\section{Results}
\label{sec:results}

\subsection{Empirical Feature-Grid Accuracy Scaling}
\label{sec:ruler}

We first evaluate displacement accuracy under the matched feature-grid design
of Section~\ref{sec:disp-design}. Particle radius, particle density,
displacement range, and raw input dimensions are scaled with the encoder
downsampling factor $\dsf{}$ such that the s2, s4, and s8 arms operate on the
same $16^3$ \featvol{} grid and encounter matched particle geometry, particle
density, and displacement magnitudes in feature-grid coordinates. The principal
difference among the three operating points is therefore the raw-volume spacing
represented by one feature-grid voxel.

\begin{table*}[t]
\centering
\caption{Displacement error under the matched feature-grid design of
Section~\ref{sec:disp-design}. Values are reported as mean $\pm$ s.d.\ over
$n=50$ independent test volumes for each arm. $EPE_{\mathrm{feature}}$ and
$EPE_{\mathrm{raw}}$ represent the same displacement error expressed in
feature-volume and raw-volume voxels, respectively, with
$EPE_{\mathrm{feature}}=EPE_{\mathrm{raw}}/\dsf{}$. Results correspond to the
single-seed released models; multi-seed training replicates are reported
separately in the text. All models use the coordinate-corrected correlation
sampler described in Section~\ref{sec:sampler}.}
\label{tab:ruler}
\begin{tabular}{cccccc}
\toprule
Arm & $\dsf{}$ & Raw input size & $\Rraw{}$ (vx)
& $EPE_{\mathrm{feature}}$ (feature vx)
& $EPE_{\mathrm{raw}}$ (vx) \\
\midrule
s2 & 2 & $32^3$  & 2 & $0.0179 \pm 0.0028$ & $0.036 \pm 0.006$ \\
s4 & 4 & $64^3$  & 4 & $0.0153 \pm 0.0036$ & $0.061 \pm 0.015$ \\
s8 & 8 & $128^3$ & 8 & $0.0166 \pm 0.0045$ & $0.133 \pm 0.036$ \\
\bottomrule
\end{tabular}
\end{table*}

\begin{figure*}[h!]
  \centering
  \includegraphics[width=0.8\textwidth]{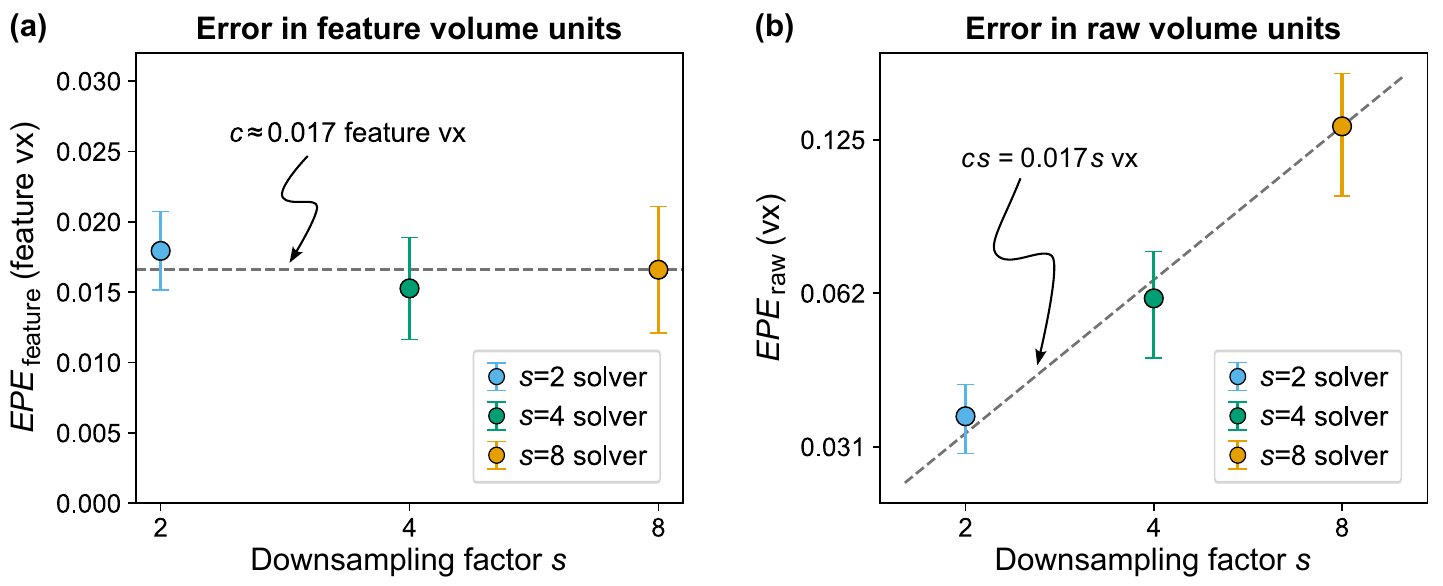}
  \caption{Empirical feature-grid accuracy scaling under the matched design of
Section~\ref{sec:disp-design}.
\textbf{(a)} Endpoint error expressed in feature-volume voxels as a function of
encoder downsampling factor $\dsf{}$. The dashed line denotes the mean over the
three arms, $c=0.017$ feature voxel.
\textbf{(b)} The same displacement errors expressed in raw-volume voxels; the
dashed line shows $c\,\dsf{}$. Points and error bars denote mean $\pm1$ s.d.\
over $n=50$ independent test volumes per arm.}
  \label{fig:ruler}
\end{figure*}

Under this matched design, the mean endpoint errors expressed in feature-volume
coordinates are $0.0179$, $0.0153$, and $0.0166$ feature voxels for s2, s4,
and s8, respectively (see Table~\ref{tab:ruler} and Fig.~\ref{fig:ruler}a). These
values cluster around a constant $EPE_{\mathrm{feature}} \approx c$  = 0.017 feature voxel.

Expressed instead in raw-volume coordinates, the corresponding mean errors are
$0.036$, $0.061$, and $0.133$ voxels for s2, s4, and s8, respectively
(Fig.~\ref{fig:ruler}b). Because one feature-volume voxel corresponds to
$\dsf{}$ raw-volume voxels, $EPE_{\mathrm{raw}} = \dsf{}\,EPE_{\mathrm{feature}}$, the approximately constant feature-space error produces the empirical scaling $EPE_{\mathrm{raw}}
 \approx c\,\dsf{}$.

Thus, across the three matched operating points tested here, the arms localize
displacement to a similar fraction of a feature-grid voxel, while the same
relative localization error corresponds to progressively larger errors when
expressed in raw-volume coordinates. An additional evaluation performed
directly on the native \featvol{} grid, before trilinear upsampling, shows the
same trend, indicating that the observed scaling is not introduced by the
final interpolation step.



\subsection{Operating Regimes of the Three Solvers}
\label{sec:solver}
Having characterized displacement accuracy under matched conditions, we next evaluate the trained solvers along two additional dimensions relevant to
deployment: displacement magnitude and particle size relative to the feature grid. These experiments quantify the displacement reach and particle-size robustness of s2, s4, and s8 outside their nominal training conditions.

\subsubsection*{Displacement Operating Range}
\begin{figure}[h!]
  \centering
  \includegraphics[width=\linewidth]{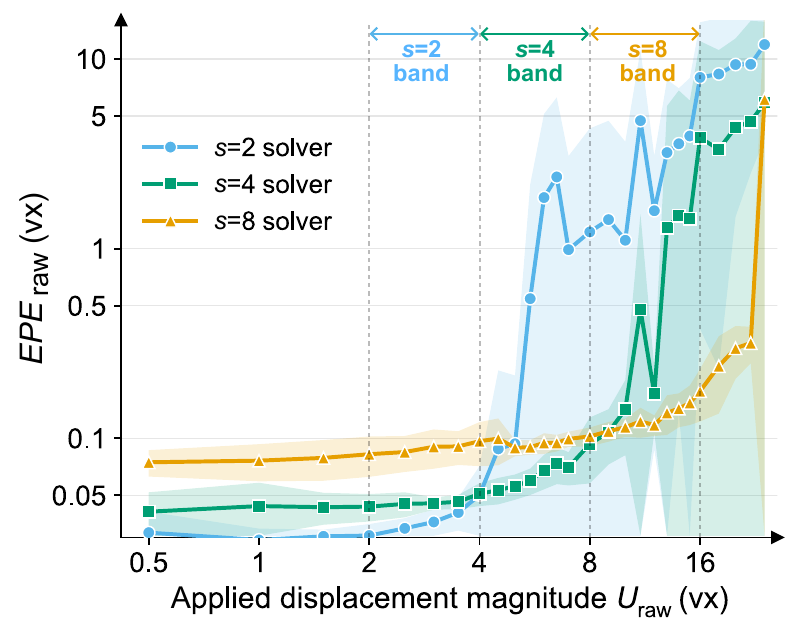}
  \caption{Displacement operating range of the three \ourmethod{} arms. \rEPE{} is plotted against the prescribed displacement amplitude
$\Uraw{}=\max_{\mathbf{x},i}|u_i(\mathbf{x})|$ for each arm evaluated at its native input size ($32^3$, $64^3$, and $128^3$ voxels for s2, s4, and s8, respectively) and matched particle morphology. Shaded vertical regions denote the corresponding training intervals $[\dsf{},2\dsf{}]$. Curves show the mean and shaded bands denote $\pm1$ s.d.\ over $n=20$ independent volumes at each displacement level. Values below the plotting limit are clipped at the logarithmic-axis floor. The accurate-range and collapse thresholds are defined in the text.}
  \label{fig:envelope}
\end{figure}

Figure~\ref{fig:envelope} shows the displacement-error response of each solver at its native input size and matched particle morphology. Each curve exhibits an approximately constant error floor over the training range, followed by a rapid loss of accuracy as the applied displacement increases.

We characterize this behavior using two operational thresholds. The
\emph{accurate range} is defined as the largest applied displacement for which
the mean \rEPE{} remains within twice the in-distribution error floor of the
corresponding arm. Using the matched errors in Table~\ref{tab:ruler}, these
threshold errors are $0.072$, $0.122$, and $0.266$ voxels for s2, s4, and s8,
respectively. The \emph{collapse point} is defined as the largest applied
displacement for which the mean \rEPE{} remains below one raw voxel.

The resulting accurate-range and collapse thresholds are $4.3$ and $5.7$
voxels for s2, $9.4$ and $12.7$ voxels for s4, and $18.9$ and $22.2$ voxels
for s8, respectively. Each solver therefore satisfies the accuracy range
criterion throughout its training interval $[\dsf{},2\dsf{}]$, with the
measured range extending beyond the upper training bound in each case.

For comparison, the geometric reach of the coarsest correlation lookup is $r\,2^{L-1}\dsf{} = 8\dsf{}$, where $r=4$ feature voxels is the correlation-lookup radius and $L=2$ is the number of correlation-pyramid levels (Section~\ref{sec:arch}). This corresponds to nominal reaches of 16, 32, and 64 raw voxels for s2, s4, and s8, respectively. These geometric scales are substantially larger than the measured operating thresholds, showing that the nominal correlation-lookup extent alone does not determine the usable displacement range of the trained solvers.

\begin{figure}[h!]
  \centering
  \includegraphics[width=\linewidth]{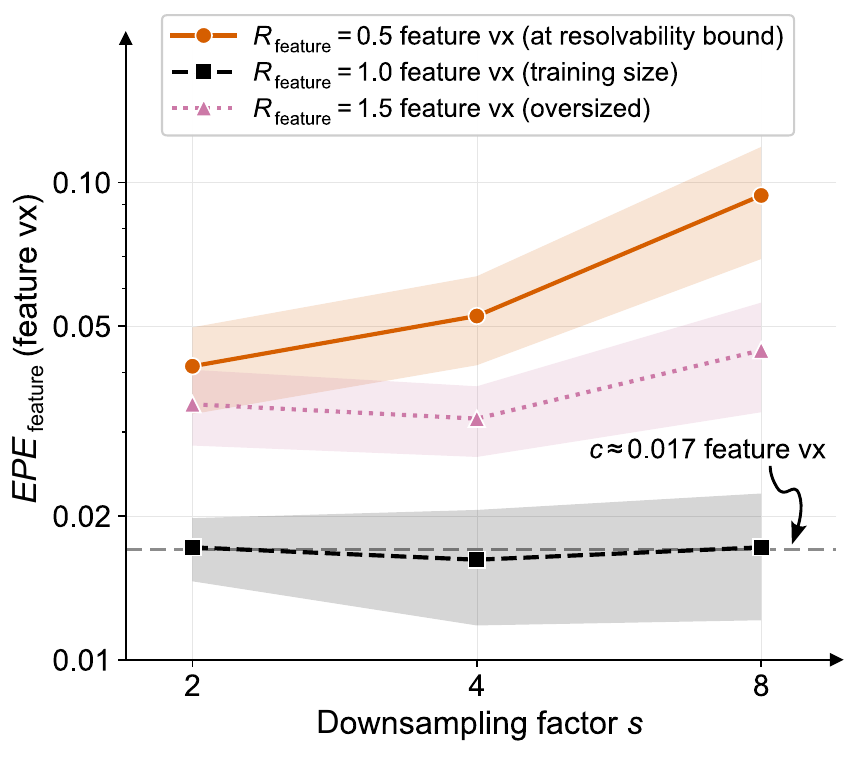}
  \caption{Sensitivity to particle size outside the training distribution.
$EPE_{\mathrm{feature}}$ is shown for feature-space particle radii
$\Rfeat{}=0.5$, $1$, and $1.5$ feature voxels across the s2, s4, and s8 arms.
For each arm, particle number density and displacement distribution are held at
their training values, so only particle radius is varied. $\Rfeat{}=1$ is the
in-distribution particle size used during training. Points and shaded bands
denote mean $\pm1$ s.d.\ over $n=30$ independent test volumes per
configuration. The dashed line denotes the matched feature-grid error scale
$c$ identified in Section~\ref{sec:ruler}.}
  \label{fig:ood}
\end{figure}

\subsubsection*{Robustness to Particle-Size Variations}
To quantify sensitivity to particle-size mismatch, we generate additional
test-only volumes at feature-space particle radii $\Rfeat{}=\Rraw{}/\dsf{}=0.5$, $1$, and $1.5$ feature voxels. For each arm, particle number density and displacement distribution are held at their training values so that particle radius is the only varied quantity. Thirty independently seeded volumes are evaluated for each configuration. The case $\Rfeat{}=1$ reproduces the particle size used during training and serves as the in-distribution reference.

All three arms achieve their lowest feature-space errors near
$\Rfeat{}=1$ (Fig.~\ref{fig:ood}). Departing from the trained particle size
increases $EPE_{\mathrm{feature}}$ by approximately two- to five-fold. For
oversized particles, $\Rfeat{}=1.5$, the error increases by approximately
$2.0$--$2.6\times$ relative to the in-distribution value. The penalty is larger
for undersized particles, $\Rfeat{}=0.5$, increasing from approximately
$2.4\times$ for s2 to $5.5\times$ for s8. The response is therefore
asymmetric with respect to particle-size mismatch, with the largest degradation
observed for particles smaller than those represented during training. These results also show that particle size relative to the feature-grid spacing is an important determinant of out-of-distribution performance.

\begin{figure}[h!]
  \centering
  \includegraphics[width=1\linewidth]{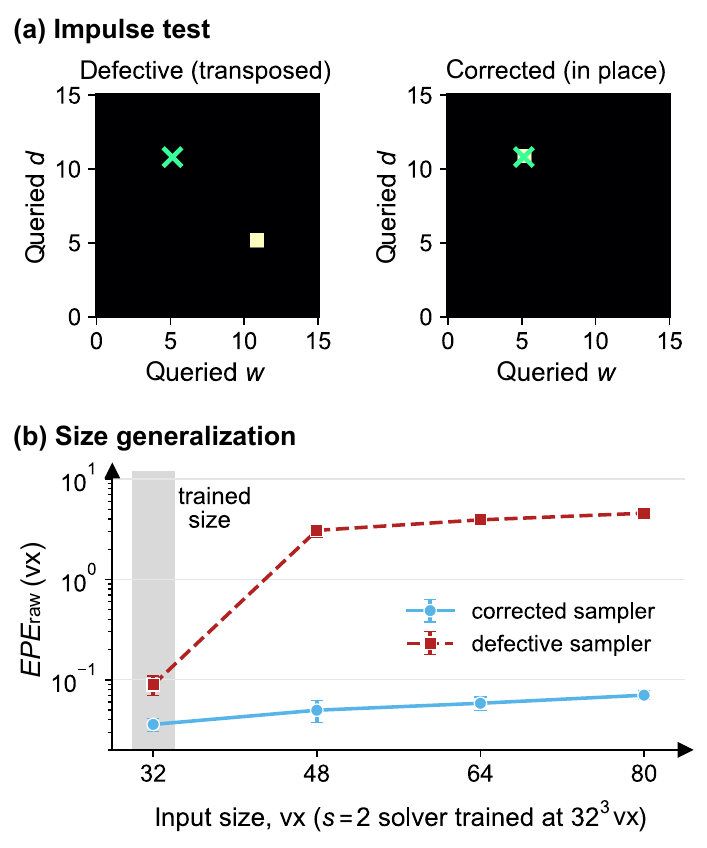}
  \caption{Correlation-sampler coordinate consistency and input-size
generalization.
\textbf{(a)} Non-cubic impulse test used to verify the sampling geometry. A
unit impulse is placed at a prescribed coordinate (green cross) and queried
through the correlation-lookup sampler. The coordinate-inconsistent sampler
queries the transposed location, whereas the corrected sampler recovers the
impulse at the prescribed coordinate.
\textbf{(b)} Input-size generalization of the s2 arm, trained on $32^3$-voxel
volumes and evaluated at progressively larger input dimensions while retaining
the same particle morphology and displacement distribution. Results are shown
for models trained and evaluated with the coordinate-inconsistent and corrected
samplers, respectively.}
  \label{fig:sizegen}
\end{figure}

\subsection{Effect of Sampler Correction and Input Size}
\label{sec:sizegen_results}

Correcting the correlation-sampling coordinate convention improves both
native-input accuracy and generalization to volume dimensions not represented
during training (Fig.~\ref{fig:sizegen}). For the s2 arm trained on
$32^3$-voxel inputs, the corrected model yields \rEPE{} values of approximately
$0.036$, $0.050$, $0.058$, and $0.070$ voxels when evaluated on $32^3$,
$48^3$, $64^3$, and $80^3$ volumes, respectively. In contrast, the otherwise
identical model trained with the coordinate-inconsistent sampler exhibits
errors $34$--$51\times$ larger when evaluated at input dimensions different
from the training size. Even at the native $32^3$ input size, correcting the
sampler reduces \rEPE{} from $0.089$ to $0.036$ voxel, corresponding to an
approximately $2.5\times$ reduction.

Tile dimensions also affect inference accuracy. For the s2 solver, the
central-region \rEPE{} is $0.029$, $0.049$, and $0.058$ voxels for tile side
lengths of 32, 48, and 64 voxels, respectively. With 50\% overlap, the mean
EPE within tile-boundary regions is approximately $10$--$17\%$ larger than in
the tile interiors. Increasing the overlap to 75\% reduces this difference to
approximately $4\%$. Thus, overlapping inference reduces but does not
completely eliminate tile-boundary effects.

\subsection{Zero-Strain Noise Floor Assessment}
\label{sec:noisefloor}

Before evaluating finite-deformation accuracy, we quantify the baseline
measurement noise of each method using zero-strain image pairs. Each pair is
generated from the same undeformed particle configuration but with independent
realizations of imaging noise. Because the prescribed displacement field is
identically zero, any recovered displacement represents measurement bias and
random error introduced by the correlation procedure.

\begin{figure}[h!]
  \centering
  \includegraphics[width=0.5\textwidth]{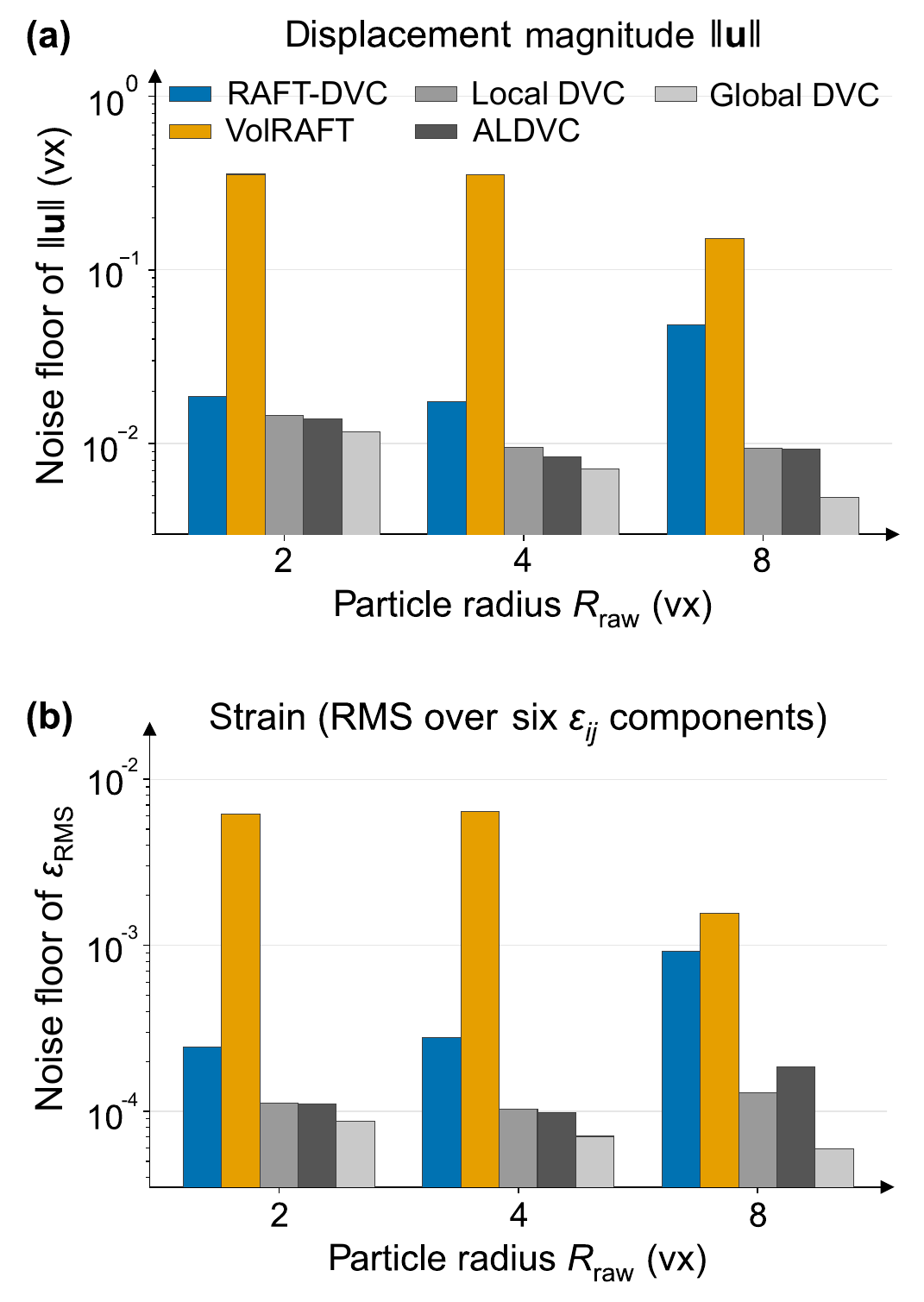}
  \caption{Zero-strain measurement noise for eight independently noise-realized static image pairs in each particle regime. \textbf{(a)} Apparent displacement magnitude $\|\mathbf{u}\|$. \textbf{(b)} RMS infinitesimal strain, $\varepsilon_{\mathrm{RMS}}$, computed from $\boldsymbol{\varepsilon}
  =1/2 \, [ \nabla\mathbf{u} + (\nabla\mathbf{u})^{\!\top} ]$ using the common gradient-estimation procedure described in the text. The prescribed displacement and strain are zero in all cases.}
  \label{fig:noisefloor}
\end{figure}

Figure~\ref{fig:noisefloor}(a) compares the displacement noise floor across the three particle regimes with different radii. The classical DVC methods, including the ICGN-based local subvolume method~\cite{aldvc}, finite-element-based global method~\cite{globaldvc}, and ALDVC method~\cite{yang2020aldvc,aldvc}, exhibit the smallest apparent
displacement errors ranging from approximately $0.005$ to $0.015$ voxels. Finite-element-based global DVC gives the lowest noise floor in all three regimes ($0.005$--$0.012$ voxels), smaller than the local and AL-DVC methods whose errors are $0.008$--$0.015$ voxels. This is because a ``$\| \nabla \mathbf{u} \|^2$'' regularization term was used in the finite-element-based global method to smooth the deformation field.

The three \ourmethod{} arms exhibit displacement noise floors of approximately $0.018$--$0.047$ voxels across the tested regimes. For example, the s2 arm gives an apparent displacement of approximately $0.019$ voxel, about $1.4\times$ the corresponding ALDVC value. The sampler-corrected VolRAFT model gives substantially larger apparent displacements, approximately
$0.15$--$0.36$ voxel.

Component-wise statistics indicate that the zero-strain error of \ourmethod{} contains limited systematic directional bias. Across the three arms, the mean bias of each displacement component remains below 0.012 voxel and is similar among the $x$, $y$, and $z$ directions. The component-wise standard deviations are approximately 0.008--0.010 voxels for the s2 and s4 arms, and about $ 0.030$ voxel for the s8 arm.  The noise floor is also found to be insensitive to imaging signal-to-noise ratio (SNR). For instance, changing the SNR from 6 to 51 only changes the displacement noise floor of s8 from $0.050$ to $0.048$ voxel, a difference below 5\%.

Figure~\ref{fig:noisefloor}(b) reports the corresponding strain noise floor. For all methods, displacement gradients are evaluated using the same local least-squares linear fit over a 32-voxel support, and the infinitesimal strain tensor is computed as
$\boldsymbol{\varepsilon}=\tfrac12(\nabla\mathbf{u}+\nabla\mathbf{u}^{\!\top})$.
The scalar strain-noise metric is defined as the root mean square of its six independent
components and averaged over all evaluation locations,
\begin{equation}
  \varepsilon_{\mathrm{RMS}}
  \;=\;\left[\frac{1}{6 N_{\mathrm{node}} }\sum_{n=1}^{N_{\mathrm{node}}}\ \sum_{i\in\mathcal{C}}
       \varepsilon_{i}^{2}(\mathbf{x}_n)\right]^{1/2},
  \label{eq:strainrms}
\end{equation}
where $\mathcal{C}=\{xx,\,yy,\,zz,\,xy,\,xz,\,yz\}$ collects the three normal and three shear components, each counted once, and $N_{\mathrm{node}}$ is the total number of evaluation locations pooled over all image pairs.

The strain-noise ordering broadly follows the displacement-noise ordering. Classical DVC yields the lowest values, approximately $0.60$--$1.9\times10^{-4}$ RMS strain. The s2 and s4 arms yield approximately $2.4$--$2.8\times10^{-4}$, whereas the s8 arm gives approximately $9.2\times10^{-4}$. The larger s8 value is because differentiation amplifies errors from the coarser feature-grid representation. VolRAFT gives substantially larger strain-noise levels, approximately $1.6$--$12.9\times10^{-3}$.

\subsection{Benchmarking against Existing DVC Methods Using Synthetic Deformations }
\label{sec:baseline}

We next benchmark \ourmethod{} against representative classical and other learning-based DVC methods using synthetic deformations. 
The benchmark is designed to evaluate four aspects of volumetric displacement measurement: (i) in-distribution accuracy, (ii) cross-regime generalization, (iii) spatial resolution of DVC analysis, and (iv) computational cost and scalability.

\subsubsection*{Benchmark scenarios and comparison protocol}
Seven benchmark scenarios, denoted S1--S7, are considered in this paper. Scenarios S1--S5 vary particle size and displacement magnitude.  Particularly, scenarios S1--S3 correspond to the three matched training regimes of \ourmethod{}, whereas S4 and S5 deliberately combine particle size and displacement magnitude from different regimes. Scenario S6 uses a frequency-swept sinusoidal displacement field to probe spatial resolution, and scenario S7 extends the coarse-particle, large-displacement condition to a $512^3$-voxel volume to assess large-volume deployment. All seven benchmark cases are summarized below:
\begin{itemize}
    \item[$\bullet$] \textbf{S1--S3: matched in-distribution (ID) cases.} 
    The particle-radius and displacement combinations are 
    $(\Rraw{},\Uraw{})=(2, [2,4])$, $(4, [4,8])$, and $(8, [8,16])$ voxels, corresponding to the s2, s4, and s8 training regimes, respectively.  Displacement fields are generated from Eq~\eqref{eq:shapefn} and scaled according to the prescribed $\Uraw{}$ range.
    
    \vspace{0.5em}
    
    \item[$\bullet$]\textbf{S4--S5: cross-regime out-of-distribution (OOD) cases.}
    S4 combines fine particles with large displacement, $(\Rraw{},\Uraw{})=(2,[8,16])$ voxels, whereas S5 combines large particles with small displacements, $(\Rraw{},\Uraw{})=(8,[2,4])$ voxels. Neither case lies within the joint particle-size and displacement distribution of a single \ourmethod{} arm.

    \vspace{0.5em}

    \item[$\bullet$] \textbf{S6: spatial-resolution probe.}
    A frequency-swept sinusoidal displacement field is prescribed such that only one displacement component is nonzero,
    \begin{equation}
      u = A\,\sin\varphi(x_2), \qquad v = w = 0 ,
      \label{eq:chirp}
    \end{equation}
    where the phase $\varphi(x_2)$ changes local spatial frequency
    \begin{equation}
        \varphi(x_2)=\varphi_0+\int_{0}^{x_2} \frac{2 \pi}{\lambda(x)}  \,\mathrm{d}x.
    \end{equation}
    
    The local wavelength $\lambda$ decreases linearly from $\lambda_0$ to $\lambda_L$ along the $x_2$ direction across the volume, and $\varphi_0$ is drawn uniformly from $[0,2\pi)$. Three particle and deformation bands are evaluated on $256^3$ volumes, with
    $(A,\lambda_0\!\rightarrow\!\lambda_L)$ equal to $(3,64\!\rightarrow\!8)$,
    $(6,96\!\rightarrow\!16)$, and $(12,128\!\rightarrow\!32)$ voxels for the fine- (s2), intermediate- (s4), and
    coarse-particle (s8) regimes, respectively. Each band uses the particle radius and number density matched to its arm under the design of Section~\ref{sec:disp-design}, that is $\Rraw{}=2$, 4, and 8 voxels at $4.8$, $0.6$, and $0.075$ particles per $10^3$ \rawvol{} voxels, so the three bands differ in texture as well as in displacement. 
    
    \vspace{0.5em}

    \item[$\bullet$] \textbf{S7: large-volume deployment case.}
    The particle size and displacement distribution  of S3 repeated on a $512^3$-voxel volume to evaluate tiled inference, memory requirements, and
    computational cost at substantially larger scale.

\end{itemize}
%

\begin{figure*}[t]
  \centering
  \includegraphics[width=0.86\textwidth]{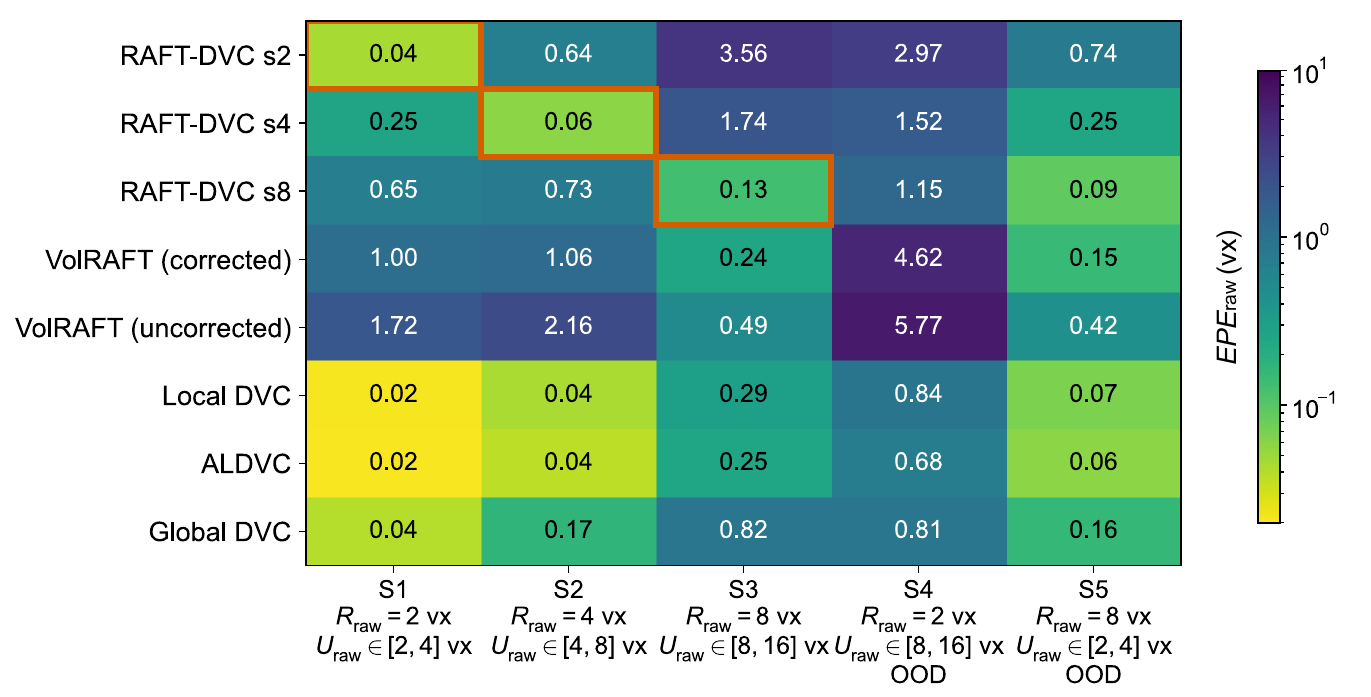}
  \caption{Displacement-accuracy benchmark across scenarios S1--S5. Cells show mean \rEPE{} in raw-volume voxels on a logarithmic color scale.
All methods are evaluated at the same $16^3=4096$ step-8 locations within the central $128^3$-voxel region, and each value is averaged over $n=20$
independent volumes. The upper rows show the three \ourmethod{} solvers and the two retrained VolRAFT variants; the lower rows show the classical DVC methods. Outlined cells ([1,1], [2,2], [3,3] locations in the matrix) identify the matched \ourmethod{} arm for each in-distribution scenario S1--S3. S4 and S5 are outside the joint training distribution of all three \ourmethod{} arms.}
  \label{fig:benchmatrix}
\end{figure*}

\begin{figure*}[tp]
  \centering
  \includegraphics[width=\textwidth]{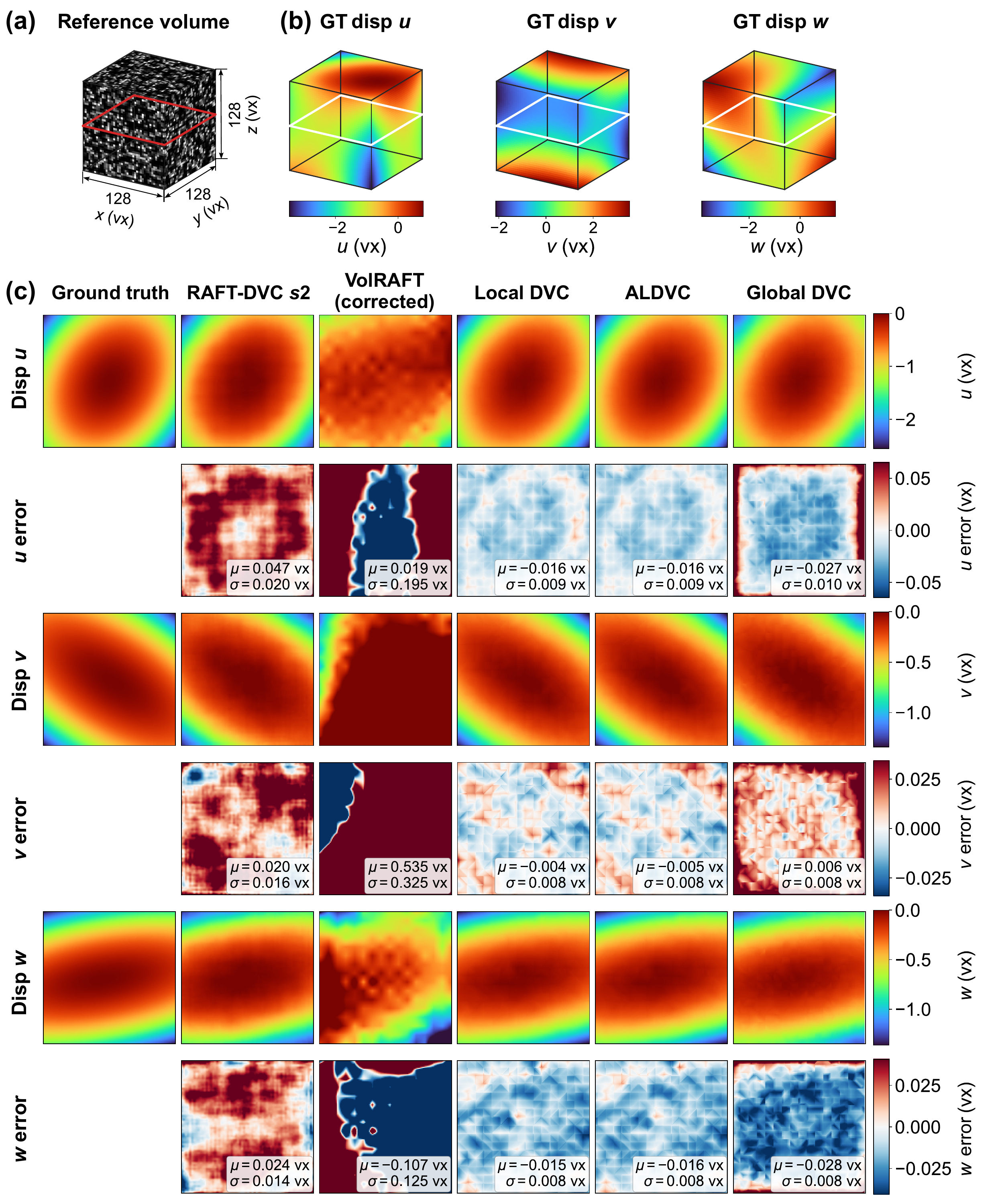}
  \caption{Representative displacement fields for the fine-particle, small-displacement benchmark S1.
(a-b) The reference particle volume and prescribed $u$, $v$, and $w$ displacement fields. The grid below shows the corresponding mid-$z$ plane
for ground truth, the matched \ourmethod{} s2 solver, sampler-corrected VolRAFT, and the three classical DVC methods. (c) Each predicted displacement component is paired with its signed error relative to ground truth, with the mean and standard deviation over the valid region annotated. Classical fields are shown only within their nodal support.} 
  \label{fig:benchfieldsA}
\end{figure*}

\begin{figure*}[tp]
  \centering
  \includegraphics[width=\textwidth]{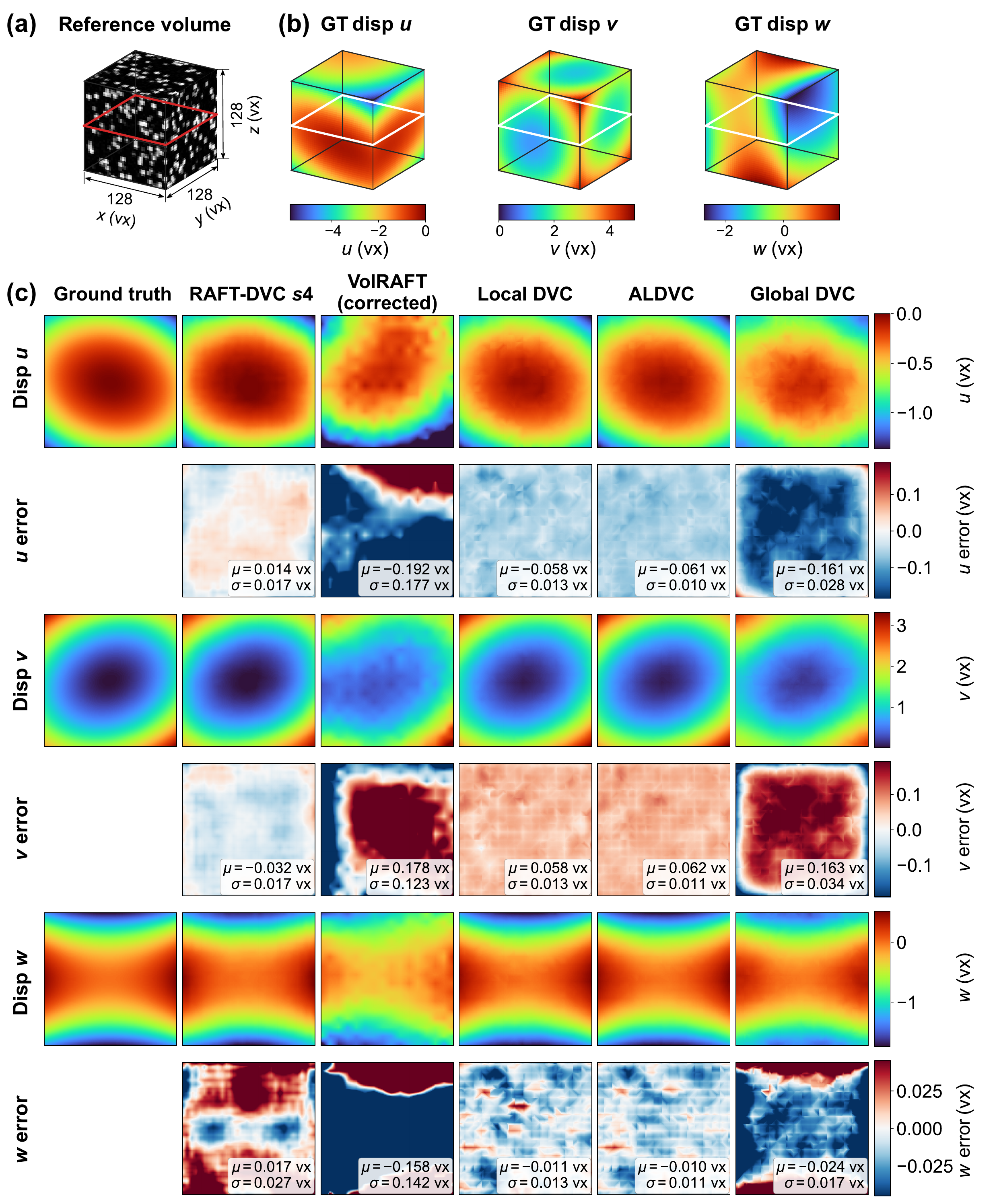}
  \caption{Representative displacement fields for the intermediate-particle benchmark S2, for which s4 is the matched \ourmethod{} solver. Layout and
notation are the same as in Fig.~\ref{fig:benchfieldsA}.} 
\end{figure*}

\begin{figure*}[tp]
  \centering
  \includegraphics[width=\textwidth]{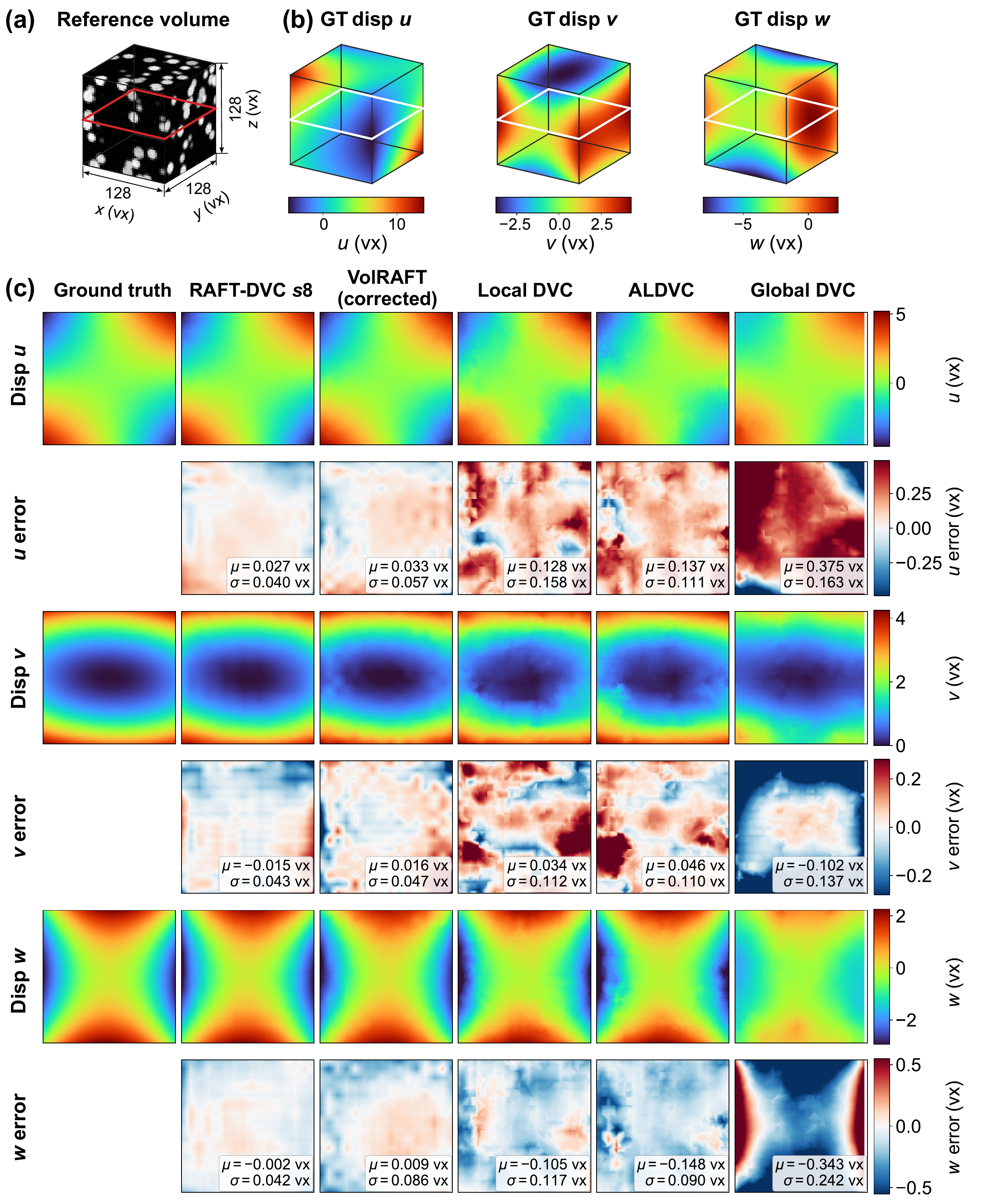}
  \caption{Representative displacement fields for the coarse-particle, large-displacement benchmark S3, for which s8 is the matched \ourmethod{}
solver. Layout and notation are the same as in Fig.~\ref{fig:benchfieldsA}.}
  \label{fig:benchfieldsC}
\end{figure*}

\begin{table*}[t]
\centering
\caption{Summary of the DVC methods evaluated in this study. \ourmethod{} comprises three resolution-aware arms, s2, s4, and s8, with encoder downsampling factors $\dsf{} = 2$, 4, and 8, respectively. VolRAFT is retrained on the same synthetic data and evaluated using both its released correlation sampler and the coordinate-corrected sampler described in
Section~\ref{sec:sampler}. The classical baselines include the local DVC method, augmented-Lagrangian DVC (ALDVC), and finite-element-based global DVC methods. The learning-based methods report dense voxel-wise displacement fields, whereas the classical methods are evaluated on nodal grids. Code and implementation details are provided in Appendix~\ref{app:disclosure}.}
\label{tab:methods}
\begin{tabular}{@{}llp{0.49\linewidth}c@{}}
\toprule
Method & Family / role & Description & $\dsf{}$ \\
\midrule
\ourmethod{}~s2 & Ours, fine-grid arm & RAFT-DVC with $\dsf{}=2$, trained on the fine-particle, small-displacement regime & 2 \\
\ourmethod{}~s4 & Ours, intermediate-grid arm & RAFT-DVC with $\dsf{}=4$, trained on the intermediate particle-size and displacement regime & 4 \\
\ourmethod{}~s8 & Ours, coarse-grid arm & RAFT-DVC with $\dsf{}=8$, trained on the coarse-particle, large-displacement regime & 8 \\
\addlinespace
VolRAFT (released sampler) & Learning-based baseline
& Published VolRAFT architecture~\cite{wong2024volraft}, retrained on our
  synthetic data using the released correlation sampler & 8 \\
VolRAFT (corrected sampler) & Learning-based baseline
& Same VolRAFT architecture and training protocol, using the coordinate-corrected correlation sampler described in Section~\ref{sec:sampler} & 8 \\
\addlinespace
Local DVC  & Classical baseline & Local subset DVC with one subset per measurement node and inverse-compositional Gauss--Newton (IC-GN) sub-voxel refinement~\cite{aldvc,yang2020aldvc} & N/A \\
ALDVC  & Classical baseline & Augmented-Lagrangian (AL) DVC coupling local subset correlation with global kinematic-compatibility constraints~\cite{aldvc,yang2020aldvc} & N/A \\
Global DVC  & Classical baseline & Finite-element-based global DVC representing the displacement field on a mesh and determining its degrees of freedom through regularized volume-wide~\cite{globaldvc} & N/A \\
\bottomrule
\end{tabular}

\end{table*}

The benchmark includes the three \ourmethod{} solvers, sampler-corrected and released-sampler VolRAFT~\cite{wong2024volraft}, and three representative classical approaches: local DVC, finite-element global DVC, and ALDVC~\cite{yang2020aldvc}. Complete classical parameter settings and the associated tuning protocol are provided in Appendix~\ref{app:disclosure}.

To isolate the effect of correlation-sampler coordinate consistency, the two VolRAFT models use the same architecture, synthetic training data, optimization procedure, training duration, and random seed; they differ only in the correlation-sampling implementation. Both models are retrained using the training protocol of Section~\ref{sec:training}. Both models also share the feature and context encoder structure, the input normalization, and the trilinear displacement upsampling of \ourmethod{}~s8; they differ from it in the correlation pyramid ($L=4$ levels at radius $r=3$ against $L=2$ at $r=4$) and in using a full rather than a separable ConvGRU.

For scenarios S1--S5, synthetic volumes are generated with a 24-voxel padding region on each face, and accuracy comparisons are performed within the common central $128^3$-voxel region. Classical DVC is applied to the complete padded volume so that subsets associated with evaluation nodes retain full image support. All methods are compared on the same step-8 evaluation grid within the central region, containing $16^3=4096$ locations. The learning-based methods produce dense displacement fields, which are sampled at these same locations for the quantitative comparison. Complete solver settings are given in Appendix~\ref{app:disclosure}.

\subsubsection*{Displacement Accuracy Within and Across Training Regimes}

Figure~\ref{fig:benchmatrix} summarizes the displacement accuracy of all methods for scenarios S1--S5. In the three matched in-distribution cases, the s2, s4, and s8 solvers achieve mean \rEPE{} values of $0.045$, $0.058$, and $0.128$ voxels in S1, S2, and S3, respectively. These values are consistent with the increase in raw-volume error with feature-grid spacing observed in Section~\ref{sec:ruler}.

For the fine- and intermediate-particle regimes S1 and S2, the tuned classical methods retain an accuracy advantage. In S1, ALDVC achieves
$\rEPE{}=0.021$ voxel, compared with $0.045$ voxel for \ourmethod{} s2. In S2, ALDVC gives $0.037$ voxel, compared with $0.058$ voxel for \ourmethod{} s4.

The ordering reverses in the coarse-particle, large-displacement case S3. Here, \ourmethod{} s8 achieves $\rEPE{}=0.128$ voxel, compared with
$0.248$ voxel for ALDVC, corresponding to an approximately $1.9\times$ reduction in endpoint error. Thus, the relative accuracy of the learning-based
and classical methods depends strongly on the particle-size and displacement regime.

Performance degrades when particle size and displacement magnitude are combined outside the joint training regimes. In S4, the best \ourmethod{} result is $\rEPE{}=1.15$ voxels, compared with $0.68$, $0.81$, and $0.84$ voxels for ALDVC, global DVC, and local DVC, respectively. In S5, the best classical result is $0.058$ voxel, whereas the s8 solver achieves $0.089$ voxel. Because neither S4 nor S5 matches the joint particle-size and displacement distribution of any \ourmethod{} arm, these scenarios constitute cross-regime out-of-distribution tests.

Figures~\ref{fig:benchfieldsA}--\ref{fig:benchfieldsC} show representative displacement components and signed error maps for the three in-distribution scenarios. The displayed samples are selected at the median \ourmethod{} EPE within each scenario.\\

\begin{figure}[h!]
  \centering
  \includegraphics[width=\linewidth]{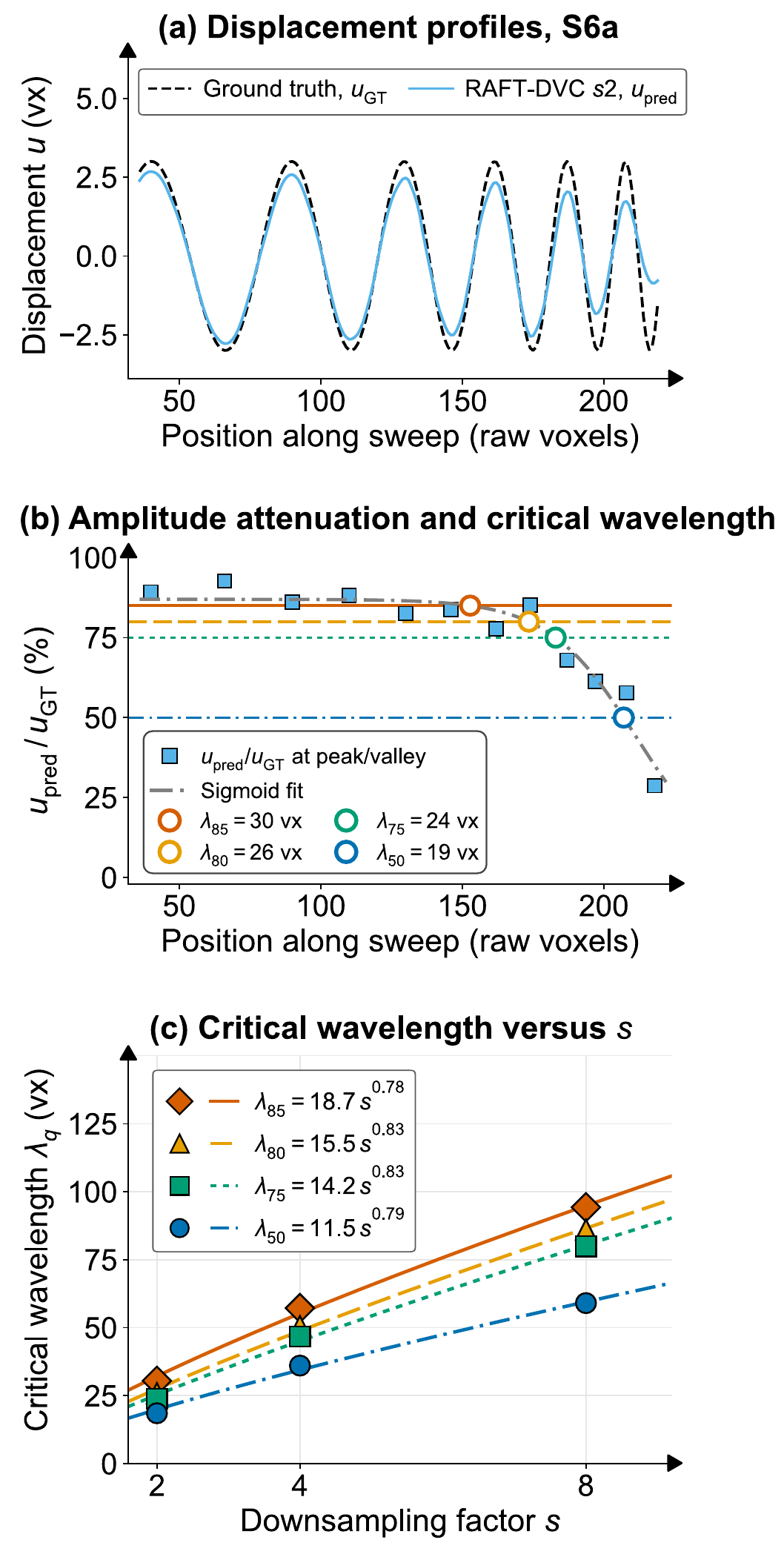}
  \caption{ Deformation spatial resolution of the matched \ourmethod{} arms in scenario S6.
\textbf{(a)} Prescribed displacement $u_\mathrm{GT}$ and displacement $u_\mathrm{pred}$ recovered by the matched s2 arm for one volume in the fine-particle band S6a. \textbf{(b)} Amplitude attenuation for the same volume, shown against the position along the sweep. Squares denote the retained amplitude fraction $u_\mathrm{pred}/u_\mathrm{GT}$ at the peaks and valleys of the displacement profile, and the dashed curve is the sigmoid fit of Eq~\eqref{eq:sigmoid}. Horizontal lines indicate retained-amplitude levels of $q=85$, $80$, $75$, and $50\%$. Open circles mark their intersections with the fitted attenuation curve; each intersection is mapped to the prescribed wavelength to obtain the corresponding critical wavelength $\lambda_q$. The fit is shown only over the position range spanned by the measured extrema. \textbf{(c)} Critical wavelength $\lambda_q$ as a function of encoder downsampling factor $\dsf{}$ for the three matched arms at the four retained-amplitude levels. Lines show power-law fits $\lambda_q=a\dsf{}^{\,b}$. Corresponding levels in (b) and (c) use the same line styles and markers. }
  \label{fig:lamcrit}
\end{figure}

\subsubsection*{Deformation Spatial Resolution}
\begin{figure*}[tp]
  \centering
  \includegraphics[width=\textwidth]{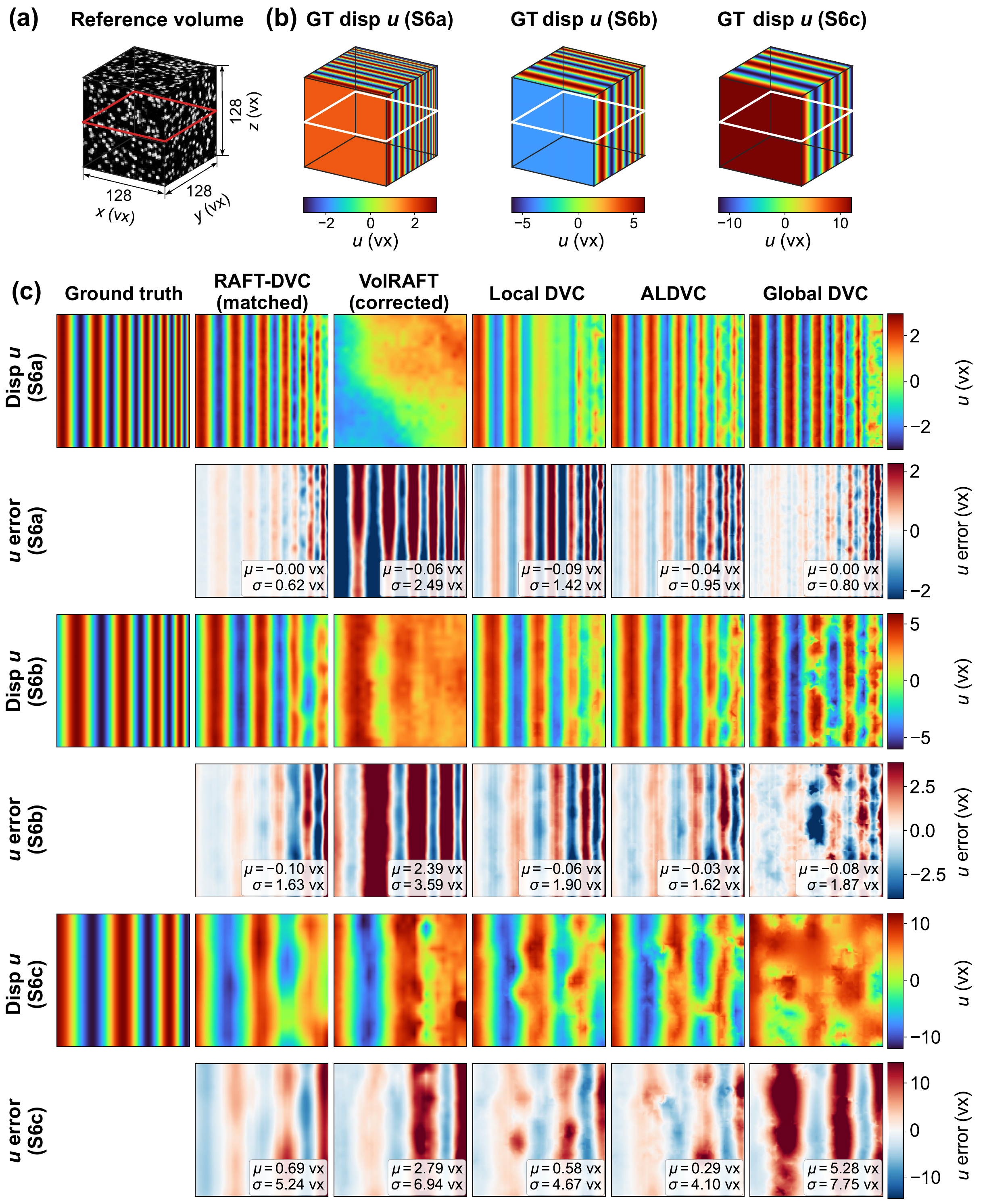}
  \caption{Spatial-resolution benchmark S6 using frequency-swept sinusoidal
displacement fields on $256^3$-voxel volumes.
The three rows correspond to the fine-, intermediate-, and coarse-particle
regimes, with displacement amplitudes $A=3$, 6, and 12 voxels and wavelength
ranges $64\!\rightarrow\!8$, $96\!\rightarrow\!16$, and
$128\!\rightarrow\!32$ voxels, respectively.
The top strip shows the reference particle volume and prescribed displacement
field for each regime. The comparison grid shows the mid-$z$ plane for ground
truth, the matched \ourmethod{} solver, sampler-corrected VolRAFT, and the three
classical DVC methods, together with signed displacement-error maps.}
  \label{fig:s6star}
\end{figure*}

\begin{figure*}[tp]
  \centering
  \includegraphics[width=0.95 \textwidth]{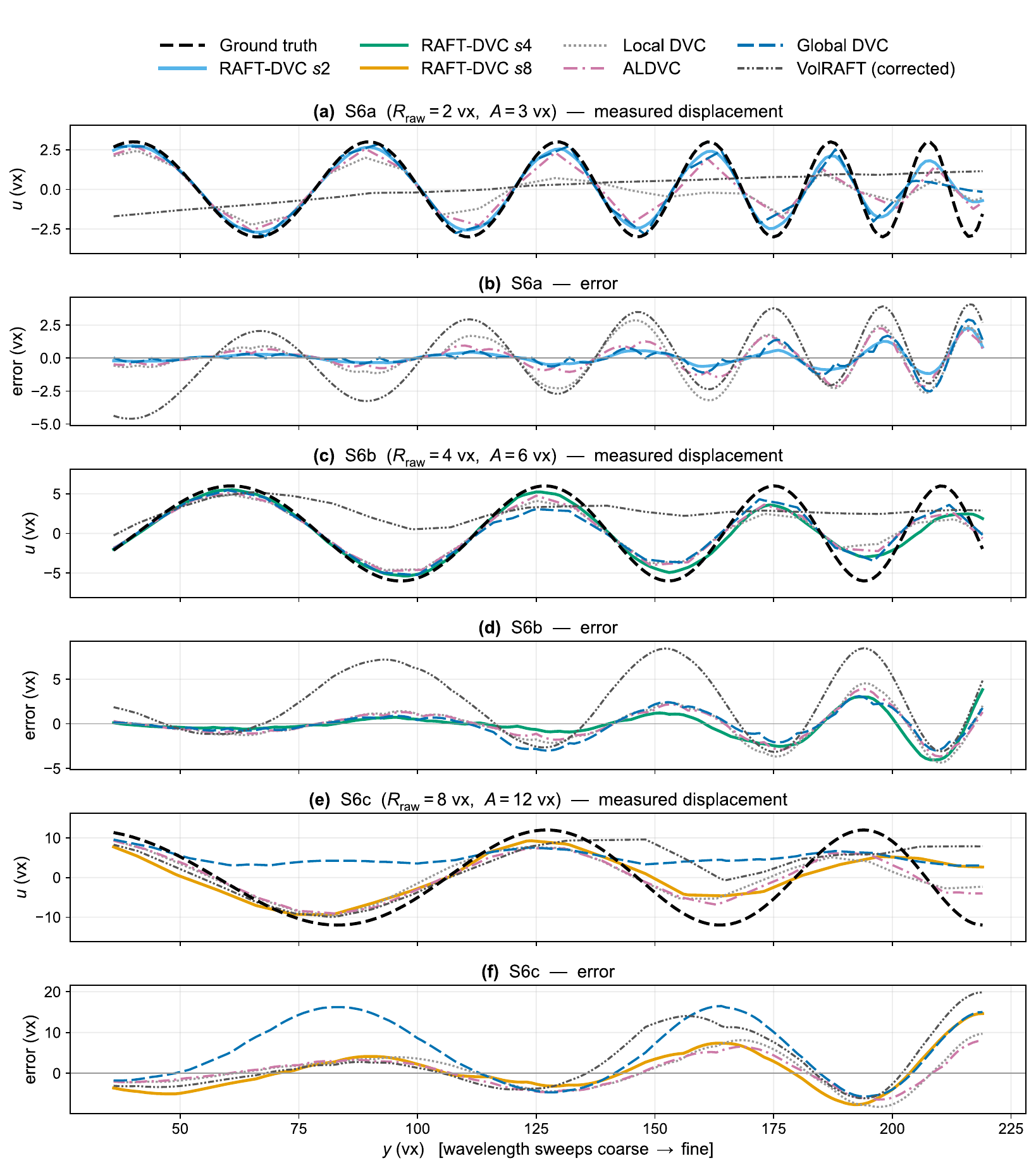}
 \caption{Displacement line profiles for the S6 spatial-resolution benchmark.
Rows correspond to the fine-, intermediate-, and coarse-particle regimes
($\Rraw{}=2$, 4, and 8 voxels). For each regime, the $u$ displacement field is
averaged across the transverse direction within the common valid region and
plotted along the wavelength-sweep direction, from coarse to fine spatial
variation. The corresponding signed displacement error is shown below each
profile. The ground-truth field, matched \ourmethod{} arm, sampler-corrected
VolRAFT, and classical DVC methods are plotted for comparison.}
  \label{fig:starcut}
\end{figure*}

Scenario S6 provides an independent measure of deformation spatial resolution using frequency-swept sinusoidal displacement fields (Figs.~\ref{fig:s6star} and~\ref{fig:starcut}). For each solver, the critical
wavelength $\lambda_q$, defined in Section~\ref{sec:metrics}, is obtained from
the attenuation curve relating recovered to prescribed displacement amplitude
(Fig.~\ref{fig:lamcrit}b). Because the initial phase of the sinusoidal sweep is
drawn independently for each volume, extrema from different realizations occur
at different spatial locations. For the intermediate- and coarse-particle
bands, extrema from three volumes are therefore pooled to increase sampling
density along the attenuation curve, whereas the fine-particle band contains
sufficient extrema to be fitted from a single volume.

For the matched solver in each band, $\lambda_{85}$ is $30$, $57$, and $94$
\rawvol{} voxels for $\dsf{}=2$, 4, and 8, respectively. The corresponding
values are $26$, $51$, and $86$ voxels for $\lambda_{80}$; $24$, $47$, and $80$
voxels for $\lambda_{75}$; and $19$, $36$, and $59$ voxels for $\lambda_{50}$
(Fig.~\ref{fig:lamcrit}c). All four threshold crossings lie within the spatial
range sampled by the measured extrema and therefore do not require
extrapolation of the fitted attenuation curve. The fitted long-wavelength
plateau $T_0$ in Eq~\eqref{eq:sigmoid} is $0.87$, $0.89$, and $0.95$ for the
three bands, indicating that the recovered amplitude remains slightly below the
prescribed amplitude even at the longest wavelengths represented in the sweep.

Under the matched design, the characteristic length scales of the synthetic
problem increase with $\dsf{}$. If deformation spatial resolution scaled
strictly in proportion to the feature-grid spacing, the critical wavelength
would satisfy $\lambda_q\propto\dsf{}$. To quantify departures from this
proportional scaling, we fit
\begin{equation}
    \lambda_q = a\,\dsf{}^{\,b},
\end{equation}
for which $b=1$ would correspond to a constant critical wavelength expressed in
\featvol{} voxels. The fitted exponents are $b=0.79\pm0.07$,
$0.83\pm0.06$, $0.83\pm0.06$, and $0.78\pm0.05$ for $q=50$, $75$, $80$,
and $85\%$, respectively. The similar exponents across attenuation thresholds
indicate a consistent sublinear dependence of deformation spatial resolution
on the encoder downsampling factor.

Accordingly, the critical wavelength increases with $\dsf{}$, but less than
proportionally. Increasing $\dsf{}$ fourfold from 2 to 8 increases
$\lambda_{85}$ from $30$ to $94$ raw-volume voxels, a factor of $3.1$, and
$\lambda_{50}$ from $19$ to $59$ voxels, a factor of $3.2$. When expressed on
the feature grid, $\lambda_{85}$ corresponds to $15.0$, $14.3$, and $11.8$
\featvol{} voxels for s2, s4, and s8, respectively. Thus, although the coarse
solver requires a larger deformation wavelength in raw-volume coordinates, it
resolves a somewhat shorter wavelength when measured relative to its own
feature-grid spacing. This behavior differs from the approximately proportional
raw-EPE scaling observed across the matched arms in
Section~\ref{sec:ruler}, showing that displacement accuracy and deformation
spatial resolution constitute distinct characteristics of solver performance.

These critical wavelengths provide a practical criterion for solver selection
under the S6 conditions. For example, the measured $\lambda_{85}$ values imply
that deformation wavelengths of approximately $30$, $57$, and $94$ raw-volume
voxels are required for s2, s4, and s8, respectively, to recover $85\%$ of the
prescribed sinusoidal amplitude. The corresponding $\lambda_{50}$ values are
$19$, $36$, and $59$ voxels. A deformation wavelength of $50$ voxels would
therefore lie above the $85\%$ threshold for s2 and above the $50\%$ threshold
for s4, but below the $50\%$ threshold for s8. Deformation spatial resolution
therefore provides a constraint on solver selection that is complementary to
displacement reach: a displacement magnitude may lie well within a solver's
operating range while its spatial variation remains too fine to be faithfully
resolved.

\begin{table*}[t]
\centering
\caption{Measured computational cost for the S3 and S7 benchmarks.
S3 uses $176^3$-voxel volumes evaluated over the central $128^3$ region,
whereas S7 uses a $512^3$ volume with a $480^3$ valid interior. Learning-based
methods are evaluated on an NVIDIA RTX~5090 GPU and return dense displacement
fields; classical methods use the reference MATLAB/CPU implementations and
return nodal displacement fields. Runtime therefore reflects differences in
hardware, implementation, output density, and nodal spacing and should not be
interpreted as a hardware-independent algorithmic speedup. Memory denotes peak
GPU allocation for the learning-based methods and peak MATLAB resident memory
for the classical implementations.}
\label{tab:benchcost}
\begin{tabular}{clcccc}
\toprule
\textbf{Scale} & \textbf{Method} & \textbf{Runtime} (s) & \textbf{Peak memory}  & \textbf{Output values} & \textbf{Output values/s} \\
\toprule
\multirow{5}{*}{S3 ($176^3$ vx)}
 & \ourmethod{} s8     & 0.12 & $0.5$~GB (GPU)   & 2.1\,M & $1.7{\times}10^7$ \\
 & VolRAFT (corrected) & 0.12 & $0.5$~GB (GPU)   & 2.1\,M & $1.8{\times}10^7$ \\
 & Local DVC          & 13.6 & $15.4$~GB (CPU)  & 4096   & 301 \\
 & ALDVC               & 28.6 & $15.4$~GB (CPU)  & 4096   & 143 \\
 & Global DVC          & 15.2 & $1.9$~GB (CPU)   & 4096   & 270 \\
\midrule
\multirow{6}{*}{S7 ($512^3$ vx)}
 & \ourmethod{} s8 ($128^3$ tiles)      & 18.4  & $0.5$~GB (GPU)  & 110.6\,M & $6.0{\times}10^6$ \\
 & \ourmethod{} s8 ($256^3$ tiless)      & 10.1  & $8.3$~GB (GPU)  & 110.6\,M & $1.1{\times}10^7$ \\
 & VolRAFT (corrected, $256^3$ tiles)   & 10.9  & $10.6$~GB (GPU) & 110.6\,M & $1.0{\times}10^7$ \\
 & Local DVC                            & 293.7 & $46.3$~GB (CPU) & 18\,954  & 65 \\
 & ALDVC                    & 363.3 & $46.3$~GB (CPU) & 18\,954  & 52 \\
 & Global DVC            & 305.6 & $11.2$~GB (CPU) & 23\,548  & 77 \\
\bottomrule
\end{tabular}
\end{table*}

\subsubsection*{Large-Volume Dataset Deployment}

Scenario S7 evaluates large-volume deployment by repeating the coarse-particle, large-displacement S3 regime on a $512^3$-voxel volume. Using overlapping $128^3$ tiles, the s8 solver predicts a dense displacement field over the $480^3$ valid interior, corresponding to $110.6$ million output values, in $18.4$~s with a measured peak GPU memory allocation of $0.5$~GB. The resulting mean \rEPE{} is $0.106$ voxel. These dense output values should
not be interpreted as independent spatial measurements: displacement is refined on the underlying feature grid and subsequently interpolated to the raw-volume grid.

Increasing the tile side length to $256^3$ reduces runtime to $10.1$~s but increases peak GPU memory to $8.3$~GB and increases \rEPE{} to $0.184$ voxel. This reproduces the tile-size dependence observed independently in Section~\ref{sec:sizegen_results}. Under the same $256^3$ tiling condition,
sampler-corrected VolRAFT requires $10.9$~s and $10.6$~GB and yields $\rEPE{}=0.370$ voxel.

The classical solvers are evaluated on coarser nodal grids at S7 to maintain practical computational cost. ALDVC, local DVC, and global DVC require approximately $5$--$7$~min on the tested CPU implementations and yield mean \rEPE{} values of $0.038$, $0.057$, and $0.127$ voxels, respectively. In comparison, the s8 solver provides an intermediate level of displacement accuracy while producing a dense raw-grid output in substantially shorter measured wall-clock time on the GPU.

These runtime comparisons should not be interpreted as hardware-independent algorithmic speedups. The learning-based and classical implementations use different hardware, software environments, output densities, and nodal
spacings. Table~\ref{tab:benchcost} therefore reports the measured computational costs as implementation-level benchmarks rather than normalized algorithmic complexity comparisons.\\

Finally, the benchmark provides an independent comparison of the released and coordinate-corrected VolRAFT samplers. In the coarse-particle, large-displacement S3 regime, \ourmethod{} s8 achieves $\rEPE{}\approx0.13$ voxel, compared with approximately $0.24$ voxel for sampler-corrected VolRAFT trained on the same synthetic regime. Across the tested benchmark cases, coordinate correction reduces the VolRAFT endpoint error by approximately $1.5$--$2.8\times$ relative to the otherwise matched released-sampler implementation. These results are consistent with the input-size tests of Section~\ref{sec:sizegen_results} and further demonstrate
the sensitivity of three-dimensional RAFT inference to correlation-sampler coordinate consistency.

\subsection{Experimental Evaluation on Confocal Volumes}
\label{sec:realdata}

The preceding benchmarks use synthetic volumes generated by a common rendering
pipeline. We therefore further evaluate the three \ourmethod{} solvers on an
experimental particle-labeled confocal indentation dataset
~\cite{tong2026dvcchallenge,yang2020aldvc}. Because no ground-truth
displacement field is available for this experiment, absolute displacement
error cannot be measured. Instead, we assess consistency with established
classical DVC solutions and complement this comparison with the reference-free
image-matching metric defined in Section~\ref{sec:metrics}.

All methods are evaluated using a common-interior protocol. Each solver receives
the same image region centered beneath the spherical indenter, with sufficient
surrounding image context for both subset-based and tiled analyses. Quantitative
comparisons are restricted to a common interior region containing approximately
2,900 classical DVC nodal locations. The immediate near-contact region is
excluded because its displacement exceeds the operating ranges characterized
for some of the trained solvers.

The confocal voxel spacings are $0.42$, $0.42$, and $0.425~\upmu$m along the
$x_1$, $x_2$, and $x_3$ directions, respectively. The imaged particles have an
equivalent diameter of approximately $5$--$6$ raw-volume voxels. Within the
common interior, the local DVC and ALDVC reference displacement fields indicate a smooth
depth-dependent deformation with a mean displacement magnitude of approximately
$5.1$ voxels and a maximum magnitude of approximately $15$ voxels. The largest
individual displacement component reaches approximately $13$ voxels.

\begin{figure*}[t]
  \centering
  \includegraphics[width=0.92\textwidth]{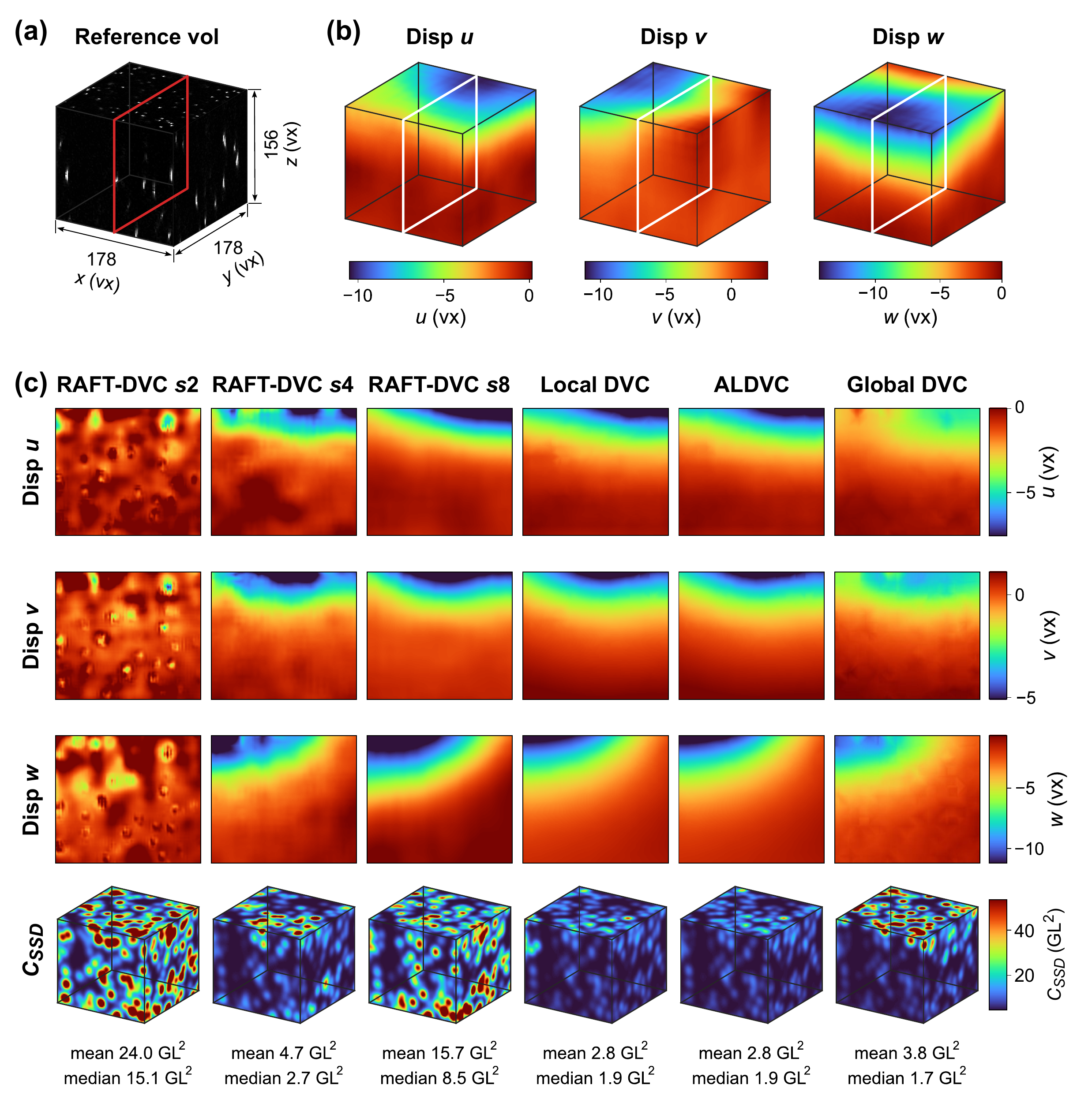}
  \caption{Experimental evaluation on a particle-labeled confocal indentation
  dataset.  \textbf{(a-b)} Reference confocal volume and the three displacement components
  predicted by the \ourmethod{} s8 solver. The spherical indenter is located
  above the top surface, and the red frame indicates the $xz$ section shown
  below.
  \textbf{(c)} Displacement components on the common $xz$ section and local mean squared intensity residual
  $C_{\mathrm{SSD}}$  for the   three \ourmethod{} solvers and the three classical DVC methods. A common
  color scale is used within each row.  }
  \label{fig:realdata}
\end{figure*}

The s4 and s8 solvers reproduce the overall spatial structure of the classical
DVC displacement field in all three components (Fig.~\ref{fig:realdata}), with
component-wise Pearson correlation coefficients $\rho$ ranging from
approximately $0.90$ to $0.99$. The s8 field reaches $\rho\approx0.99$ relative to the ALDVC reference field
and has a mean vector disagreement of approximately $1$ voxel. In comparison, the s2
solver exhibits a mean vector disagreement of approximately $3.1$ voxels over
the same region.

The larger disagreement of s2 is strongly associated with local displacement
magnitude. Its measured collapse point from Section~\ref{sec:solver} is
$5.7$ voxels, whereas the largest displacement component in the experimental
region reaches approximately $13$ voxels. Approximately $27\%$ of the evaluated
locations lie beyond the characterized s2 collapse point. The median vector
disagreement increases from approximately $0.63$ voxel where the local
displacement is below $2$ voxels, to approximately $2.6$ voxels near the
measured s2 operating limit, and to approximately $13.1$ voxels in the
largest-displacement bin. Averaged over all evaluated locations, the
disagreement is approximately $1.0$ voxel within the characterized s2 range and
$7.7$ voxels beyond it.

By comparison, fewer than $1\%$ of the evaluated locations exceed the measured
s4 collapse point of $12.7$ voxels, and none exceed the s8 collapse point of
$22.2$ voxels. The relative performance of the three solvers is therefore
consistent with the displacement operating ranges measured independently on
synthetic data.

The experimental particle morphology also differs from the matched synthetic
training conditions. An experimental particle radius of approximately
$2.5$--$3$ raw voxels corresponds to feature-space radii of approximately
$1.25$--$1.5$, $0.63$--$0.75$, and $0.31$--$0.38$ feature voxels for s2,
s4, and s8, respectively. Thus, none of the three solvers encounters exactly
the particle scale used during training, $\Rfeat{}=1$, and the experimental
particles are particularly small relative to the s8 feature-grid spacing. The
interaction between this texture mismatch and displacement reach is considered
further in Section~\ref{sec:deployment}. \\

\begin{table}[t]
\centering
\caption{Reference-free image-matching quality for the experimental confocal
dataset. $C_{\mathrm{SSD}}$ is the local mean squared intensity residual
defined in Section~\ref{sec:metrics}, evaluated over the common interior using
$\sigma_d=1$ voxel and a $w=21$-voxel local window. Residual values are
reported in squared gray levels (GL$^2$). The null displacement field provides
a no-registration baseline. Mean displacement magnitude is reported only to
characterize the displacement field returned by each method.}
\label{tab:sir}
\footnotesize
\begin{tabular}{lcccc}
\toprule
 & & \multicolumn{3}{c}{$C_{\mathrm{SSD}}$ (GL$^2$)} \\
\cmidrule(l){3-5}
Method & Mean $\|\mathbf{u}\|$ (vx) & Mean & Median & $p_{90}$ \\
\midrule
Null field ($\mathbf{u}\equiv\mathbf{0}$)
    & 0    & 46.1 & 0.81 & 9.46 \\
\midrule
\ourmethod{} s2
    & 2.75 & 23.8 & 0.26 & 3.40 \\
\ourmethod{} s4
    & 4.44 & 5.0  & 0.12 & 1.59 \\
\ourmethod{} s8
    & 4.69 & 15.8 & 0.16 & 2.60 \\
\midrule
Local DVC
    & 4.94 & 2.9 & 0.11 & 1.28 \\
ALDVC
    & 4.95 & 2.8 & 0.11 & 1.27 \\
Global DVC
    & 4.44 & 4.6 & 0.12 & 1.56 \\
\bottomrule
\end{tabular}
\end{table}

Agreement with a classical DVC solution does not establish absolute accuracy, because the methods could in principle share a common error. We therefore also evaluate the reference-free squared intensity residual defined in Section~\ref{sec:metrics}, which measures how well the estimated displacement field aligns the deformed volume with the reference volume. 

The null displacement field produces a mean residual of $46.1$~GL$^2$ (Table~\ref{tab:sir}). The s2 estimate reduces this value to $23.8$~GL$^2$ but remains substantially above the residuals obtained with the other non-null displacement fields. The s4 solver gives a mean residual of $5.0$~GL$^2$, comparable to the $4.6$~GL$^2$ obtained with global finite element-based DVC. Local DVC and ALDVC yield the smallest mean residuals, $2.9$ and $2.8$~GL$^2$, respectively. The s8 solver produces a mean residual of $15.8$~GL$^2$ despite its high component-wise correlation with the local DVC and ALDVC reference displacement field.

The residual therefore provides information complementary to global displacement-field agreement and reveals differences in local image-matching quality that are not captured by the correlation coefficient alone. The residual field is also spatially heterogeneous. For s8, the mean residual varies by approximately a factor of $3.7$ across four depth bands within the evaluated region, whereas the corresponding variation for the classical reference is approximately $1.9\times$. These spatial variations further illustrate the value of the residual field as a local, reference-free measure of image-registration quality.

Finally, we note that for computation of $C_{\mathrm{SSD}}$, the dense \ourmethod{} fields are
available directly at every raw-volume voxel, whereas the classical methods
return displacements on a 12-voxel nodal grid. The local DVC and ALDVC reference displacement
fields are therefore interpolated to the raw-volume grid before image warping.
We use the same trilinear interpolation implied by their underlying
piecewise-linear displacement representations (as we use an affine shape function in local and ALDVC methods; and a linear Q8 element in the finite-element-based global DVC method). This interpolation constitutes
an additional reconstruction step for the classical methods but does not alter
their original nodal displacement estimates.

\section{Discussion}
\label{sec:discussion}

\subsection{Feature-Grid Resolution as a Measurement-Design Variable}
A key finding of this study is that, across the three matched operating points, displacement error remains approximately constant when expressed in
feature-grid coordinates,  (i.e., $EPE_{\text{feature}} \approx c$), whereas the same error expressed in raw-volume coordinates scales approximately with the encoder downsampling factor $s$ (i.e., $EPE_{\text{raw}} \approx cs$). For the present architecture and matched test conditions, $c\approx0.017$ feature voxel. This empirical relationship provides a simple geometric interpretation of the accuracy--resolution trade-off: if different solvers localize displacement to a similar fraction of one feature-grid voxel, then reducing the raw-volume spacing represented by that voxel directly reduces the corresponding error in measurement coordinates.

The matched design was constructed to make this comparison meaningful by scaling particle radius, particle density, displacement magnitude, and raw-volume dimensions together so that the correlation and recurrent-update
modules encounter comparable problems in feature-grid coordinates. This is more controlled than exchanging encoder downsampling factors while holding the raw-volume particle field fixed, because the latter would simultaneously alter particle morphology and displacement magnitude as represented on the feature grid.

The empirically observed scaling relation still needs to be investigated to determine whether it is a universal law, and to investigate how coefficient $c$ may depend on image texture, signal-to-noise ratio, learned feature representation, recurrent architecture, correlation implementation, and training distribution. The matched experiment also compares complete trained arms rather than isolating feature-grid spacing from every other aspect of representation learning. 

The comparison with sampler-corrected VolRAFT further clarifies the scope of this scaling. In S3, both \ourmethod{} s8 and the corrected VolRAFT model use the same encoder downsampling factor, $\dsf{}=8$, yet their \rEPE{} values are approximately $0.13$ and $0.24$ voxel, respectively. The two networks are close in architecture: they share the feature and context encoder structure, the joint input normalization, the trilinear displacement upsampling, and the number of recurrent iterations, and they are trained on the same data with the same optimization schedule. They differ in how the correlation volume is queried, $L=2$ pyramid levels at radius $r=4$ against $L=4$ levels at $r=3$, so that VolRAFT spans a longer nominal reach but samples the correlation peak more coarsely, and in the recurrent operator, which is separable in \ourmethod{} and full in VolRAFT. Thus, the downsampling factor determines how an error measured in feature-grid units is converted to raw-volume units, but it does not determine how accurately a particular network can localize displacement on that feature grid.

\subsection{Joint Dependence on Texture and Displacement}

The operating-regime experiments (see Section~\ref{sec:solver}) show that feature-grid spacing cannot be
selected from displacement magnitude alone. Solver performance depends jointly
on two distinct requirements: the displacement must remain within a recoverable
range, and the volumetric texture must remain compatible with the representation
learned by the solver.

\begin{figure}[h!]
  \centering
  \includegraphics[width=\linewidth]{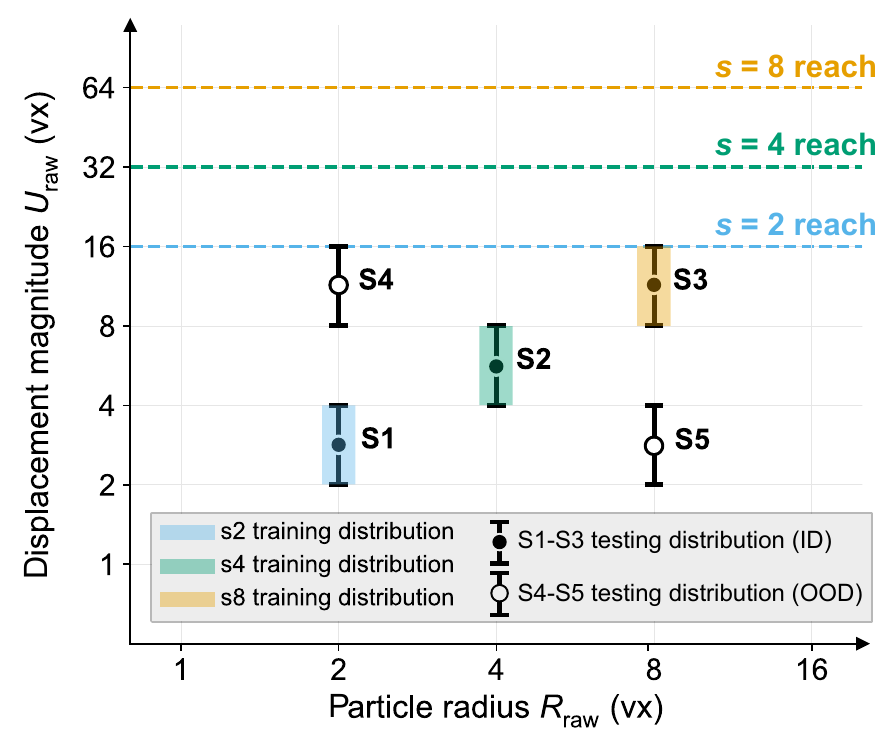}
  \caption{Benchmark scenarios S1--S5 in the
$(\Rraw{},\Uraw{})$ plane. Both quantities are expressed in raw-volume voxels.
Colored bands denote the particle-size and displacement distributions used to
train the s2, s4, and s8 arms,
$\Rraw{}=\dsf{}$ and $\Uraw{}\in[\dsf{},2\dsf{}]$.
Black segments denote the displacement intervals sampled in each benchmark
scenario. S1--S3 reproduce the corresponding matched training regimes, whereas
S4 and S5 combine particle size and displacement magnitude from different
regimes. Dashed lines indicate the nominal architectural correlation-lookup
reach $8\dsf{}$, corresponding to 16, 32, and 64 raw voxels for s2, s4, and
s8, respectively.  }
\label{fig:selmap}
\end{figure}

Figure~\ref{fig:selmap} summarizes these two constraints and plots scenarios S1--S5 on the ($\Rraw{}$, $\Uraw{}$) plane. The colored bands on S1--S3 denote the joint particle-size and
displacement distributions used to train the s2, s4, and s8 arms:
$\Rraw{}=2$, 4, and 8 voxels with
$\Uraw{}\in[2,4]$, $[4,8]$, and $[8,16]$ voxels, respectively. The nominal
architectural correlation-lookup reach is $r\,2^{L-1}\dsf{}=8\dsf{}$, where $r=4$ feature voxels is the lookup radius and $L=2$ is the number of
correlation-pyramid levels. These nominal reaches correspond to 16, 32, and
64 raw voxels for s2, s4, and s8, respectively. They describe the geometric
extent accessible to the correlation lookup and should not be interpreted as
the validated displacement range of the trained solvers.

The matched benchmark cases S1--S3 lie within the joint training distributions
of s2, s4, and s8, respectively. S4 and S5 instead combine texture and
displacement characteristics drawn from different regimes and therefore expose
two complementary forms of cross-regime mismatch.

S4 combines the fine particles characteristic of s2 with the large
displacements characteristic of s8. The s2 solver encounters familiar particle
morphology, but the imposed displacement lies well beyond its measured
operating range. Conversely, s8 has substantially greater measured displacement
reach in the present training design, but the same $\Rraw{}=2$-voxel particles
correspond to $\Rfeat{}=2/8=0.25$ feature voxel, far below the particle scale represented during s8 training.
Thus, none of the three narrowly trained solvers matches both dimensions of the
S4 condition. Correspondingly, the best \ourmethod{} result in S4 has
$\rEPE{}\approx1.15$ voxels, whereas the three tuned classical DVC methods
remain below one voxel.

S5 provides the complementary example. Its displacement magnitude is within
the range represented during s2 training, but its $\Rraw{}=8$-voxel particles
are far outside the s2 training texture. The s8 solver achieves
$\rEPE{}=0.089$ voxel, compared with $0.742$ voxel for s2, even though the
S5 displacement magnitude is substantially smaller than that used to train s8.
This result emphasizes that displacement magnitude alone is not sufficient for
solver selection.

Taken together, S4 and S5 indicate that the operating regime of a pretrained
DVC solver is inherently multidimensional. In the present solver family,
successful deployment requires both adequate displacement reach and compatible
volumetric texture. A solver that satisfies only one of these conditions can
still fail substantially outside its joint training distribution.

\subsection{Learning Relocates Rather Than Eliminates DVC Parameter Selection}
Classical DVC requires the user to choose analysis parameters such as subvolume
or element size, search range, nodal spacing, regularization strength, and
convergence settings, often adapting them to the image texture and deformation
regime of each experiment. A pretrained learning-based solver can remove much
of this continuous per-measurement tuning, but it does not eliminate the
underlying measurement-design problem. Instead, part of that choice is shifted
from selecting numerical parameters to selecting an appropriate trained solver
and operating regime.

No single \ourmethod{} arm is optimal over all particle-size and displacement
conditions examined here. Smaller $\dsf{}$ produces a finer internal grid and,
under matched conditions, lower displacement error in raw-volume coordinates,
but for a fixed raw-volume size it incurs a rapidly increasing correlation
memory cost. The solvers trained at larger $\dsf{}$ in the present design cover
larger measured displacement ranges, although this reach depends strongly on
the displacement distribution represented during training rather than on
downsampling alone. At the same time, increasing $\dsf{}$ makes a particle of
fixed raw size occupy fewer feature-grid voxels and can move the image texture
outside the morphology represented during training.

The practical question is therefore not simply ``which encoder resolution is
best?'', but rather ``which trained solver has a characterized operating regime
compatible with the texture, displacement scale, and computational constraints
of the measurement?'' In this sense, learning does not eliminate the
parameter-selection problem in DVC, but instead relocates it from continuous
per-measurement tuning toward the selection of a trained solver with a
characterized operating regime. Measurements falling outside all available
regimes may require a different solver or retraining on a broader distribution.

\subsection{Relationship to Classical DVC Methods}
The two operating constraints identified above, texture compatibility and displacement reach, have close analogues in classical DVC, although the underlying mechanisms are not identical. In subset-based correlation, measurement precision depends on the image information available within the subset. Insufficient or poorly resolved texture degrades the conditioning of
the correlation problem and increases displacement uncertainty. Large displacements pose a different challenge because successful convergence depends on the search strategy, initialization, and the basin of attraction of the iterative solution. Multiscale strategies are therefore commonly used to extend the recoverable displacement range.

The learned solvers examined here exhibit an analogous separation between texture compatibility and displacement reach. Particle features must remain sufficiently represented after feature extraction for the correlation volume to retain useful localization information, while the displacement must remain within a range represented adequately by the correlation lookup and the
training distribution. The analogy should not be taken as a one-to-one equivalence, but it provides a useful interpretation of why these two measurement constraints remain distinct even after the correlation procedure is learned.

The benchmark results also show that \ourmethod{} should not be interpreted as a universal accuracy replacement for classical DVC. In the fine- and intermediate-texture, small-to-moderate-displacement cases S1 and S2,
\ourmethod{} achieves displacement errors of the same order as the tuned classical methods, although the latter retain a modest per-measurement accuracy advantage. In the coarse-particle, large-displacement regime S3, however, the ordering reverses: the matched s8 solver reduces EPE by approximately $1.9\times$ relative to ALDVC. The relative performance of learned and classical correlation is therefore itself operating-regime dependent.

Learning-based inference also introduces a different computational and output-sampling trade-off. Once an appropriately trained solver has been selected, a dense displacement field can be inferred without the per-volume iterative parameter adjustment required by conventional DVC workflows. For the implementations benchmarked here, this produces substantially lower measured wall-clock times and enables dense inference on large volumetric datasets. These timing differences should not, however, be interpreted as hardware-independent algorithmic speedups because the learned and classical implementations use different hardware, software environments, numerical strategies, and output grids.

Though learning-based methods generate voxelwise dense deformation fields more efficiently, dense raw-grid output should not be equated with independent spatial resolution. The frequency-sweep experiment of S6 makes this quantitative. On a $256^3$ volume, the s8 solver returns a displacement vector at each of
$1.7\times10^{7}$ raw voxels, yet recovers only $80\%$ of the prescribed amplitude at a deformation wavelength of $93$ voxels (Fig.~\ref{fig:lamcrit}); its output grid is about ninety times finer along each axis than the finest deformation it reproduces at that level. The classical solvers oversample their own limit as well, but by a far smaller factor: in the intermediate-particle band ALDVC reports on an $8$-voxel node grid against $\lambda_{80}=65$ voxels, an oversampling of about eight per axis, whereas the matched s4 arm reports at every
voxel against $\lambda_{80}=57$ voxels, an oversampling of about sixty. Dense output is thus a difference in how the recovered field is sampled, not in how much independent information the measurement contains.

Though voxelwise fields produced by \ourmethod{} do not imply voxel-scale spatial resolution, the matched s2, s4, and s8 solvers retain deformation spatial resolution comparable to or better than ALDVC while offering dense inference, substantially reduced per-measurement parameter adjustment, and favorable displacement accuracy over their respective operating regimes.

\subsection{Generalization and Implementation Robustness}

The correlation-sampler study highlights a requirement that is particularly important for learning-based volumetric metrology: geometric implementation correctness should be verified independently of apparent network performance. A coordinate-order inconsistency in a three-dimensional correlation lookup can remain difficult to detect when both training and testing use fixed-size cubic volumes. The tensor dimensions remain compatible, and a network may partially adapt to the resulting representation, allowing apparently reasonable performance to coexist with an incorrect coordinate convention.

The non-cubic impulse test provides a simple architecture-independent diagnostic for this failure mode. Because the spatial dimensions are unequal, an axis permutation becomes directly observable and can be tested before any network is trained. The substantial improvement in both native-input accuracy and input-size generalization after sampler correction demonstrates that good performance at the training volume size alone is not sufficient evidence of a geometrically consistent implementation.

After coordinate correction, memory becomes the principal practical constraint on direct large-volume inference in the present architecture. The three-dimensional all-pairs correlation volume scales as the sixth power of
feature-grid side length, making fine feature grids increasingly expensive as the raw volume grows. Overlapping tiled inference alleviates this constraint by making peak memory depend primarily on tile size rather than complete volume size.

The tiled experiments nevertheless show that architectural compatibility with arbitrary spatial dimensions does not imply statistical invariance to input size. Accuracy changes as tile dimensions depart from those represented during training, and overlapping boundaries introduce an additional localized error.
Thus, input dimension and tiling configuration should be treated as part of the characterized operating regime of a deployed solver rather than as purely implementation-level details.

\subsection{Experimental Deployment}
\label{sec:deployment}

The confocal indentation experiment provides a useful test of how the synthetically characterized operating regimes translate to measured data for which the true displacement field is unknown. In this setting, solver selection must be guided by measurable properties of the images and by consistency among independent validation criteria rather than by ground-truth EPE.

The displacement-range analysis provides a clear explanation for the poor performance of s2 in the high-deformation portion of the experiment. Approximately $27\%$ of the evaluated locations lie beyond the measured s2 collapse point, compared with fewer than $1\%$ for s4 and none for s8. The s2 disagreement with the classical reference field is correspondingly strongly dependent on local displacement magnitude: the mean disagreement is approximately $1.0$ voxel within the characterized s2 range and increases to $7.7$ voxels beyond it. This correspondence between the synthetic operating range and the spatial distribution of experimental disagreement supports the practical relevance of characterizing displacement reach before deployment.

Texture compatibility gives a less direct prediction. The experimental particles have a radius of approximately $2.5$--$3$ raw voxels, corresponding to $\Rfeat{}\approx1.25$--$1.5$, $0.63$--$0.75$, and $0.31$--$0.38$ feature
voxels for s2, s4, and s8, respectively. Thus, none of the solvers operates exactly at its trained feature-space particle scale, and the mismatch is particularly large for s8. Nevertheless, s8 shows the strongest global
component-wise correlation with the classical reference field. The experimental result therefore reinforces that particle scale alone is not sufficient to select a solver; texture compatibility and displacement reach must be considered jointly.

The reference-free intensity residual adds a complementary criterion. The null field gives a mean $C_{\mathrm{SSD}}$ of $46.1$~GL$^2$, whereas s2, s4, and s8 reduce this value to $23.8$, $5.0$, and $15.8$~GL$^2$, respectively. The s4 residual is close to the $4.6$~GL$^2$ obtained with global DVC, while s8
retains a substantially larger local residual despite its high global displacement-field correlation. Its residual also varies by approximately a factor of $3.7$ across depth bands.

These observations illustrate why no single scalar criterion is sufficient for experimental solver selection. Displacement-range consistency identifies the s2 failure, global field agreement favors s8, and local image-registration quality favors s4 over s8. In the absence of ground truth, the operating-range map, cross-method displacement agreement, and reference-free image residual therefore provide complementary evidence rather than interchangeable measures of accuracy.

\subsection{Limitations and Future Work}

Several limitations define the scope of the present conclusions. First, the controlled scaling, operating-range, and benchmarking experiments use particle-labeled volumes generated by a single synthetic rendering pipeline. Although the confocal indentation experiment provides an independent experimental demonstration, it represents only one imaging modality and one deformation configuration, and no independently known ground-truth displacement field is available. Evaluation across additional volumetric imaging modalities, materials, particle morphologies, deformation modes, and image-quality conditions is therefore needed.

\begin{figure*}[h!]
  \centering
  \includegraphics[width=0.95\linewidth]{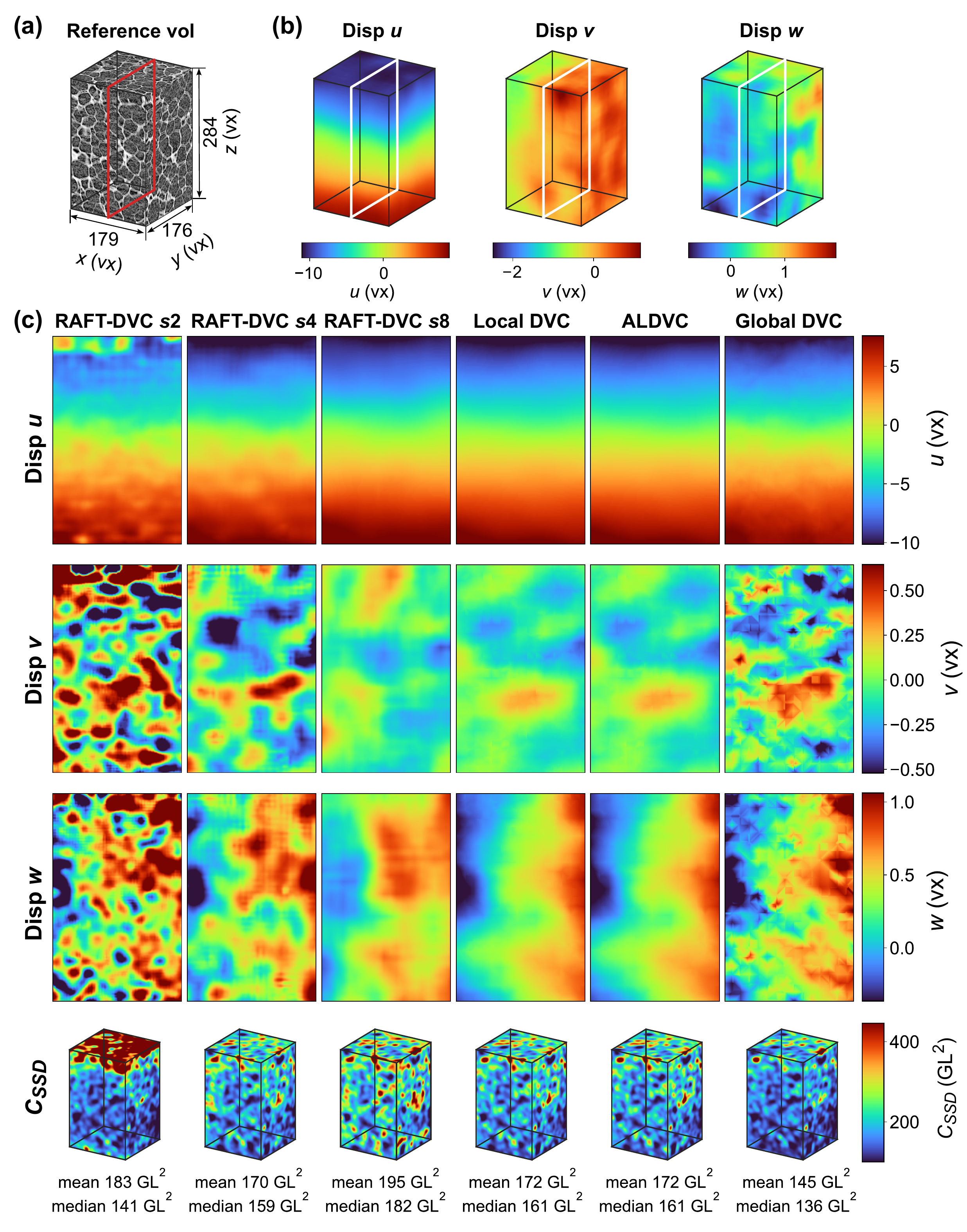}
  \caption{Generalization to a non-particle texture: elastomeric foam under
  uniaxial compression, imaged by micro-computed tomography~\cite{landauer_materials_2023,landauer_foam_dataset_2023}.
  A rigid translation of $60$ voxels was removed from the deformed volume before any
  solver was run, so the displacements shown are the deviation from rigid
  motion. \textbf{(a-b)} Reference volume and the three displacement components
  estimated by the \ourmethod{} s8 solver. The compression axis is vertical in
  the displayed volume and $u$ is the axial component; the red frame indicates
  the $xz$ section shown below. \textbf{(c)} Displacement components on the
  common $xz$ section and local mean squared intensity residual
  $C_{\mathrm{SSD}}$ for the three \ourmethod{} solvers and the three classical
  DVC methods. The \ourmethod{} solvers are applied without retraining;
  none was trained on a continuous cellular texture.}
  \label{fig:foam}
\end{figure*}

Evidence that the learned solvers can transfer beyond particle-labeled textures is provided by Fig.~\ref{fig:foam}, where the same three arms are applied without retraining to micro-computed-tomography volumes of an elastomeric foam under uniaxial compression~\cite{landauer_materials_2023,landauer_foam_dataset_2023}. The foam is a dense cellular solid whose texture consists of continuous cell walls rather than isolated features, a class absent from the training distribution and from every other experiment reported here. Over the common evaluation region the s4 and s8 arms track the ALDVC solution to $0.42$ and $0.45$ voxel, whereas the s2 arm departs from them precisely where its measured operating range predicts. Solvers trained only on synthetic particle volumes therefore transfer to a different material and a different imaging modality with a quantified rather than assumed loss of agreement, and extending the training distribution to non-particle textures is a natural next step.

Second, the matched design intentionally scales encoder downsampling together with particle radius, particle density, displacement magnitude, and raw input dimensions. It therefore characterizes three complete trained solver regimes rather than isolating the causal contribution of encoder downsampling alone. Future studies that independently vary architecture, feature-grid spacing, and training distribution would help separate these effects and determine which aspects of the observed feature-grid scaling generalize across network families.

Third, the present out-of-distribution analysis focuses primarily on particle size and displacement magnitude. Particle density, contrast, anisotropy, signal-to-noise ratio, depth-dependent image degradation, and more general descriptors of volumetric texture should be varied independently to establish a broader criterion for feature-space resolvability. Likewise, displacement reach may be extended through broader, multiscale, or curriculum-based training rather than by changing feature-grid spacing alone.

Fourth, the reference-free intensity residual used for experimental evaluation measures image-registration quality rather than displacement accuracy. It can therefore complement, but not replace, ground-truth validation. Experimental datasets with independently measured or analytically constrained deformation
fields would provide a stronger basis for evaluating absolute learned-DVC accuracy.

Finally, the memory cost of three-dimensional all-pairs correlation remains substantial, particularly for fine feature grids. Architectures that preserve a fine internal representation while avoiding the $n^6$ memory scaling of dense all-pairs correlation may provide a more favorable combination of raw-volume accuracy, displacement reach, spatial resolution, and large-volume scalability.

\section{Conclusion}
\label{sec:conclusion}

We developed \ourmethod{}, a resolution-aware family of RAFT-based DVC solvers with encoder downsampling factors $\dsf{}=2$, 4, and 8, and systematically characterized their displacement accuracy and operating regimes for volumetric images. Under a matched feature-grid design, the three arms localize displacement to approximately the same fraction of a feature voxel, leading to the empirical scaling $ \rEPE{} \approx c\dsf{}$, $ c\approx0.017$ feature voxel.
This result identifies feature-grid spacing as an important measurement-design variable that sets the raw-volume accuracy scale of the learned solver, while the achievable localization accuracy on the feature grid remains dependent on the network architecture, learned representation, and training procedure.

The three solvers exhibit complementary operating regimes rather than a single universal ranking. Finer feature grids provide lower raw-volume displacement error under matched conditions, whereas the solvers trained at coarser resolutions in the present design cover larger displacement ranges and reduce the memory cost of all-pairs correlation for a fixed raw-volume size. Performance outside the training regimes further shows that solver selection must account jointly for displacement reach and volumetric texture compatibility. In comparisons with classical DVC, tuned classical methods retain an accuracy advantage in the fine- and intermediate-particle, small-to-moderate-displacement regimes, whereas \ourmethod{} becomes competitive
or advantageous in the coarse-particle, large-displacement regime. The learned solvers also provide dense volumetric displacement fields with substantially lower measured wall-clock time in the implementations benchmarked here. Experimental evaluation on confocal indentation data further illustrates that
displacement reach, texture compatibility, cross-method field agreement, and reference-free image-matching quality provide complementary criteria for selecting a trained solver when ground truth is unavailable.

We additionally identified coordinate-order inconsistencies that can arise in three-dimensional RAFT correlation sampling and introduced a simple non-cubic impulse test for verifying sampling geometry independently of network training. Correcting the sampler substantially improves both native-input accuracy and
generalization to volume dimensions not represented during training, highlighting the importance of implementation-level geometric validation in learning-based volumetric metrology.

Taken together, these results show that learning does not eliminate the parameter-selection problem in DVC, but instead relocates it from continuous per-measurement tuning toward the selection of a trained solver with a
characterized operating regime. By releasing the three trained models, the correlation-sampling diagnostic, and the synthetic-data generation pipeline, we aim to provide a reproducible foundation for fast, dense, and quantitatively characterized learning-based DVC.

\acknowledgement{J.Y.\ gratefully acknowledges support from the U.S. National
Science Foundation (NSF) under Grants No.\ 2441460 and 2452029.  J.Y.\ and L.B.\
acknowledge the 2024--2025 Academic Development Fund from the Cockrell School of
Engineering at the University of Texas at Austin. Z.T.\ thanks the University
Graduate Continuing Fellowship. Z.T.\ and L.B.\ also acknowledge the Graduate
Excellence Fellowship and the Professional Development Award from the Cockrell
School of Engineering at the University of Texas at Austin. The authors acknowledge
the Texas Advanced Computing Center (TACC) at The University of Texas at Austin for
providing HPC resources (award DDM26002, on the Vista and Lonestar6 systems) that
have contributed to the research results reported within this paper.}

\section*{Author contributions}
Z.T.\ and J.Y.\ conceived and coordinated the RAFT-DVC project. J.Y.\ supervised
the overall research. Z.T., L.B., and J.Y.\ designed the deep-learning
methodology, the network architecture, and the training strategy. Z.T.\
implemented the RAFT-DVC framework and trained the networks. Z.T.\ performed the
quantitative analysis and benchmarked RAFT-DVC against conventional DVC methods. Z.T.\ wrote the original manuscript. All authors contributed to the interpretation of results and to the preparation and editing of the manuscript. All authors read and approved the final manuscript.

\section*{Declarations}
\paragraph{Funding.} As stated in the Acknowledgements.

\paragraph{Conflict of interest.}
The classical DVC baselines used in this study, including local subset DVC,
finite-element global DVC, and ALDVC, were evaluated using a MATLAB
implementation~\cite{yang2020aldvc} developed and maintained by the authors'
group; J.Y.\ is an author of this software. This relationship is disclosed
because the baseline implementations were configured and evaluated by authors
with direct expertise in their use, and the comparison was therefore not
performed blindly. To facilitate independent assessment and reproducibility,
all baseline parameters are reported in Appendix~\ref{app:disclosure}, and the
corresponding headless evaluation drivers are released with the accompanying
code.
Z.T.\ and J.Y.\ are also co-authors of the DVC Challenge~2.0
dataset~\cite{tong2026dvcchallenge,tong2026dvcchallengedata}, from which the
experimental confocal volumetric images acquired during indentation were
obtained. The dataset is publicly available and was used without modification
for the present study. The authors declare no financial or other competing
interests.

\paragraph{Ethics approval.} Not applicable. This study uses synthetic volumes
and a previously published, publicly available experimental dataset; it involves
no human participants or animals.

\paragraph{Data and code availability.} See the statement in
Appendix~\ref{app:disclosure}.

\appendix
\section{Reproducibility and Implementation Details}
\label{app:disclosure}

\begin{table*}[t]
\caption{Classical-DVC parameters used for each evaluation scenario. Each row lists the settings used by the three classical solvers, and $n$ denotes the number of reference--deformed volume pairs averaged for that condition. All lengths are reported in \rawvol{} voxels. Subset size is scaled with particle radius; Local DVC and ALDVC use the same subset because Local DVC corresponds to the subset-level correlation problem used within ALDVC. For finite element-based (FE-) global DVC, the element side length equals the nodal spacing. Local DVC and ALDVC use first-order affine subset shape functions, whereas FE-based global DVC uses trilinear
hexahedral elements. The regularization weight $\alpha$ is a numerical solver parameter rather than a measured quantity. Additional implementation details and solver settings are provided below.}
\label{tab:classicalparams}
\centering\footnotesize
\setlength{\tabcolsep}{5pt}
\begin{tabular}{@{}lcc@{\hskip 14pt}cc@{\hskip 14pt}cc@{\hskip 14pt}cc@{}}
\toprule
 & & & \multicolumn{2}{c}{Local DVC} & \multicolumn{2}{c}{ALDVC}
   & \multicolumn{2}{c}{FE-based Global DVC} \\
\cmidrule(lr){4-5}\cmidrule(lr){6-7}\cmidrule(l){8-9}
Test & Volume (vx) & $n$
 & Subset size (vx) & Step (vx)
 & Subset size(vx) & Step (vx)
 & Element size(vx)   & $\alpha$ \\ 
\midrule
Benchmark S1 & $176^3$ & 20 & 13 & 8 & 13 & 8 & 8   & 8 \\
Benchmark S2 & $176^3$ & 20 & 21 & 8 & 21 & 8 & 8   & 8 \\
Benchmark S3 & $176^3$ & 20 & 33 & 8 & 33 & 8 & 8   & 8 \\
Benchmark S4 & $176^3$ & 20 & 13 & 8 & 13 & 8 & 8   & 8 \\
Benchmark S5 & $176^3$ & 20 & 33 & 8 & 33 & 8 & 8   & 8 \\
\addlinespace
Benchmark S6a & $256^3$ & 3 & 13 & 8 & 13 & 8 & 8   & 0.8 \\
Benchmark S6b & $256^3$ & 3 & 21 & 8 & 21 & 8 & 8   & 0.8 \\
Benchmark S6c & $256^3$ & 3 & 33 & 8 & 33 & 8 & 8   & 0.8 \\
\addlinespace
Zero-strain (fine)   & $176^3$ & 8 & 13 & 8 & 13 & 8 & 8  & 8 \\
Zero-strain (medium) & $176^3$ & 8 & 21 & 8 & 21 & 8 & 8 & 8 \\
Zero-strain (coarse) & $176^3$ & 8 & 33 & 8 & 33 & 8 & 8  & 8 \\
\addlinespace
Benchmark S7 & $512^3$ & 3 & 33 & 16 & 33 & 16 & 16   & 8 \\
Exp indentation & $216{\times}244^2$ & 1 & 53 & 12 & 53 & 12 & 12  & 8 \\
\bottomrule
\end{tabular}
\end{table*}

\subsection{Hardware and software.} 
The three \ourmethod{} arms ($\dsf{}=2$, 4, and 8), together with the primary
evaluation and benchmark calculations, were trained and evaluated on an NVIDIA
GeForce RTX~5090 GPU (32\,GB) under Windows~11 using PyTorch nightly
\texttt{2.11.0.dev20260211} with CUDA~12.8. Training used automatic mixed
precision (fp16), while the all-pairs correlation volume and correlation lookup
were computed in fp32. Reported inference times were measured with the GPU
otherwise idle. The multi-seed robustness replicates in
Section~\ref{sec:ruler} were trained on a TACC Vista node equipped with an NVIDIA
GH200 GPU. Classical-DVC calculations were performed in MATLAB~R2025b on an
Intel Core Ultra~9 285K CPU using a 12-worker parallel pool.

\subsection{Computational cost.}
Per-arm training time and memory usage, together with the scaling of inference
correlation memory, are reported in Table~\ref{tab:cost}
(Section~\ref{sec:cost}). All three arms use the same optimization schedule of
$300$ epochs with $250$ optimization steps per epoch. Under the matched design,
the arms differ in raw-volume size and encoder downsampling factor, which
together determine their computational requirements. Each training
configuration contains $2000$ training volumes and $200$ validation volumes
generated deterministically using the parameters described below. Each trained
checkpoint is approximately $8.5$\,MB.

\subsection{Network and optimization.}
The three arms share the same overall RAFT-DVC framework, recurrent update
architecture, channel dimensions, and optimization procedure, while their
encoder variants differ in the placement of stride-2 operations to produce
downsampling factors $\dsf{}=2$, 4, and 8. The feature encoder produces
$128$-channel feature volumes and uses instance normalization. The recurrent
update maintains a $96$-channel hidden state and a $64$-channel context
representation, uses separable three-dimensional convolutions, and performs
$12$ refinement iterations. The correlation lookup uses a two-level pyramid ($L=2$)
with radius $r=4$, corresponding to a nominal geometric lookup extent of
$r2^{L-1}=8$ feature voxels, or $16$, $32$, and $64$ \rawvol{} voxels for
$\dsf{}=2$, 4, and 8, respectively. These values describe the architectural
lookup extent rather than the empirically validated displacement operating
range.

Training uses AdamW with weight decay $5\times10^{-5}$, gradient clipping at
$1.0$, and the RAFT sequence loss with temporal discount $\gamma=0.8$. A
one-cycle learning-rate schedule is used with a peak learning rate of
$2\times10^{-4}$, a warm-up fraction of $0.2$, and cosine annealing. Training
runs for $300$ epochs at an effective batch size of $8$, using gradient
accumulation where required, with base random seed $42$.

\subsection{Synthetic-data parameters.}

In addition to the rendering procedure described in Section~\ref{sec:data},
synthetic volumes contain spherical particles with a one-voxel linearly tapered
surface and a point-spread-function width of $\sigma=0.8$ voxel. Particle
centers are sampled uniformly, and overlapping particle intensities are
combined by saturation. Shot noise is modeled as Poisson noise with gain $500$,
and read noise as additive Gaussian noise with standard deviation
$\sigma=0.01$ relative to normalized intensity. Independent noise realizations
are generated for the reference and deformed volumes. Random seeding is
deterministic from a configuration hash with base seed $20260416$.

The particle radius, number density, and \rawvol{} size of the three arms follow
the matched design of Section~\ref{sec:disp-design}: radii $2/4/8$ voxels,
densities $4.8/0.6/0.075$ particles per $10^3$ raw voxels, and volume sizes
$32^3/64^3/128^3$ voxels, respectively. These choices correspond to a common
feature-grid particle radius and a common feature-grid displacement range of
$[1, 2]$ voxels.

Each displacement field is generated by drawing the shape-function coefficients
and globally rescaling the resulting field such that its largest displacement
\emph{component} equals the sampled target value. Thus, the prescribed
$\Uraw{}$, and hence each training interval $[\dsf{},2\dsf{}]$, is defined using
the componentwise maximum ($\ell_\infty$ convention). The maximum displacement
vector magnitude in the same volume can therefore exceed $\Uraw{}$ by as much
as $\sqrt{3}$ and is approximately $1.2\,\Uraw{}$ on average for the generated
fields. Displacement errors, in contrast, are reported as Euclidean vector
magnitudes throughout the paper (Section~\ref{sec:metrics}); only the
prescribed-displacement axis uses the componentwise convention.

\subsection{Evaluation metrics.}

\rEPE{} denotes the mean endpoint error expressed in \rawvol{} voxels, whereas $ EPE_{\mathrm{feature}}
= {\rEPE{}}/{\dsf{}} $ expresses the same displacement error in feature-grid voxel units. This
normalization is used when comparing the three arms under the matched design.
Out-of-distribution robustness in Fig.~\ref{fig:ood} is reported as
$EPE_{\mathrm{feature}}$ as particle size is varied at each arm's training
density; the complete particle-radius--density evaluation grid is released with
the accompanying code. Benchmark displacement accuracy is evaluated at the
common comparison nodes within the valid interior region against the prescribed
reference displacement field.

\subsection{Classical DVC methods parameters.} 

Table~\ref{tab:classicalparams} reports the principal parameters used for each
classical-DVC evaluation.

For synthetic tests S1--S6, the nodal spacing is fixed at $8$ voxels so that the three classical methods and the learned displacement fields sampled at the
same locations are compared on a common grid. The spacing is increased to
$16$ voxels for the $512^3$ deployment-scale test S7, for which an $8$-voxel
spacing would require approximately $2.3\times10^5$ subset evaluations per
volume. A spacing of $12$ voxels is used for the experimental indentation
volume, whose particle texture is substantially sparser.

Subset size is scaled with particle size following conventional DVC practice.
For particle radii $\Rraw{}=2$, 4, and 8 voxels, subset side lengths of $13$,
$21$, and $33$ voxels are used, corresponding to approximately
$4$--$6.5$ particle radii. Subset dimensions are reported using the conventional
odd-side notation $2r+1$, centered on the evaluation node. The released
implementation specifies the same window using the corresponding even
parameter $2r$; for example, the $13$-voxel subset reported here is entered as
$12$ in the solver configuration. For finite element-based (FE-based) Global DVC, the element side length
is set equal to the nodal spacing so that the finite-element and subset-based
methods are evaluated at comparable spatial sampling intervals.

ALDVC uses FFT-based integer-displacement initialization, outlier removal with a
correlation threshold of $1.25$ and median-filter threshold of $2$, and an
inverse-compositional Gauss--Newton (IC-GN) solver with tolerance $10^{-2}$ and
a maximum of $50$ iterations. The global ALDVC update uses two ADMM outer
iterations with $\mu=10^{-3}$ and
$\beta=10^{-2}\,\times\,[\mathrm{element \ size}]^{2}\,\times\mu$, together with a finite-difference
regularization subproblem. Local DVC consists of the corresponding subset-level
IC-GN problem without the global compatibility update. Both methods use
first-order affine subset shape functions with $12$ degrees of freedom per
subset.

FE-based Global DVC uses trilinear hexahedral elements defined on the nodal grid,
second-order Gauss quadrature, seven-point-stencil image gradients, the same
FFT-based integer-displacement initialization, and a maximum of $20$ iterations
with convergence tolerance $10^{-2}$. Its objective supplements the image
correlation residual with the regularization term
$\alpha\lVert\nabla\mathbf{u}\rVert^2$. The parameter $\alpha$ therefore
controls the trade-off between fidelity to local image information and spatial
smoothness of the recovered displacement field. We use $\alpha=8$ for all
tests except the frequency-sweep resolution probe S6, for which
$\alpha=0.8$ is used. The weaker regularization in S6 limits artificial
attenuation of the short-wavelength displacement variations that the test is
specifically designed to quantify.

All classical-DVC calculations were performed in MATLAB~R2025b using a
12-worker parallel pool on an Intel Core Ultra~9 285K CPU.

\subsection{Code and data availability.} 

The \ourmethod{} implementation, trained checkpoints for the three arms, the
non-cubic impulse test used to identify the correlation-sampler coordinate-order
inconsistency, headless evaluation drivers for the classical baselines, and the
complete synthetic-data generator are available at
\url{https://github.com/zachtong/RAFT-DVC} (release \texttt{v1.0}) and archived
on Zenodo at \url{https://doi.org/10.5281/zenodo.21578267}. The trained
checkpoints are included with the archived release. Synthetic benchmark volumes
are not redistributed because they can be regenerated deterministically using
the released generator and the parameters reported in this appendix. The
experimental confocal volumetric image stacks acquired during indentation are
part of the publicly available DVC Challenge~2.0
dataset~\cite{tong2026dvcchallenge,tong2026dvcchallengedata}.

\bibliographystyle{spmpsci}
\bibliography{references}

\end{document}